\RequirePackage{fix-cm}
\documentclass[twocolumn]{svjour3}          
\smartqed  
\usepackage{graphicx}
\usepackage{siunitx}
\usepackage{booktabs}
\usepackage{todonotes}
\usepackage{amsmath} 
\usepackage{amssymb}  
\usepackage{algorithm2e} 
\usepackage{multirow} 

\newcommand{\etalcite}[2]{\cite{#2}}

\usepackage[round]{natbib}

\newcommand{\hide}[1]{}

\newcommand{\ie}{{i.e.,~}}
\newcommand{\eg}{{e.g.,~}}

\newcommand{\bdmath}{\begin{dmath}}
\newcommand{\edmath}{\end{dmath}}
\newcommand{\beq}{\begin{equation}}
\newcommand{\eeq}{\end{equation}}
\newcommand{\bdm}{\begin{displaymath}}
\newcommand{\edm}{\end{displaymath}}
\newcommand{\bea}{\begin{eqnarray}}
\newcommand{\eea}{\end{eqnarray}}
\newcommand{\beal}{\beq \begin{array}{ll}}
\newcommand{\eeal}{\end{array} \eeq}
\newcommand{\beas}{\begin{eqnarray*}}
\newcommand{\eeas}{\end{eqnarray*}}
\newcommand{\ba}{\begin{array}}
\newcommand{\ea}{\end{array}}
\newcommand{\bit}{\begin{itemize}}
\newcommand{\eit}{\end{itemize}}
\newcommand{\ben}{\begin{enumerate}}
\newcommand{\een}{\end{enumerate}}

\newcommand{\Real}{\mathbb{R}}

\newcommand{\calC}{{\cal C}}

\newcommand{\calS}{{\cal S}}

\newcommand{\at}[1]{^{(#1)}}

\newcommand{\World}{\mathtt{W}}

\newcommand{\Base}{\mathtt{{B}}}

\newcommand{\meters}{\rm{m}}

\usepackage{newtxtext,newtxmath}
\journalname{Autonomous Robots}
\begin{document}

\title{Navigating by Touch: Haptic Monte Carlo Localization via Geometric Sensing and Terrain Classification 
}

\titlerunning{Navigating by Touch}        

\author{Russell Buchanan         \and
        Jakub Bednarek \and
        Marco Camurri \and
        Micha\l~R.~Nowicki \and
        Krzysztof Walas \and
        Maurice Fallon
}


\institute{R. Buchanan \and M. Camurri \and M. Fallon \at
              Oxford Robotics Institute, University of Oxford, UK \\
              \email{\{russell,mcamurri,mfallon\}@robots.ox.ac.uk}          
           \and
           J. Bednarek \and M. Nowicki \and K. Walas \at
Institute of Robotics and Machine Intelligence, Poznan University of Technology, Poland\\             \email{\{name.surname\}@put.poznan.pl}
}

\date{Received: date / Accepted: date}

\maketitle

\begin{abstract}
Legged robot navigation in extreme environments can hinder the use of
cameras and laser scanners due to darkness, air obfuscation or sensor damage. In
these conditions, proprioceptive sensing will continue to work reliably. In this paper, we propose a purely proprioceptive localization algorithm which fuses information from both geometry and terrain class, to localize a legged robot within a prior map. First, a terrain classifier computes the probability that a foot has stepped on a particular terrain class from sensed foot forces. Then, a Monte Carlo-based estimator fuses this terrain class probability with the geometric information of the foot contact points. Results are demonstrated showing this approach operating online and onboard a ANYmal B300 quadruped robot traversing a series of terrain courses with different geometries and terrain types over more than \SI{1.2}{\kilo\meter}. The method keeps the localization error below \SI{20}{\centi\meter} using only the information coming from the feet, IMU, and joints of the quadruped.
\end{abstract}

\section{Introduction}
\label{sec:intro}

Recent advances in the maturity and robustness of quadrupedal robots have made them appealing
for dull and dirty industrial operations such as routine inspection and
monitoring. Automating these operations in underground mines and sewers is particularly challenging due to darkness, in-air dust, dirt and water vapor, which can significantly impair a robot's vision system. Additionally, camera or laser sensor failure may leave only proprioceptive sensors (\ie IMU and joint encoders) at the robot's disposal.

\begin{figure}
 \centering
 \includegraphics[width=0.95\columnwidth]{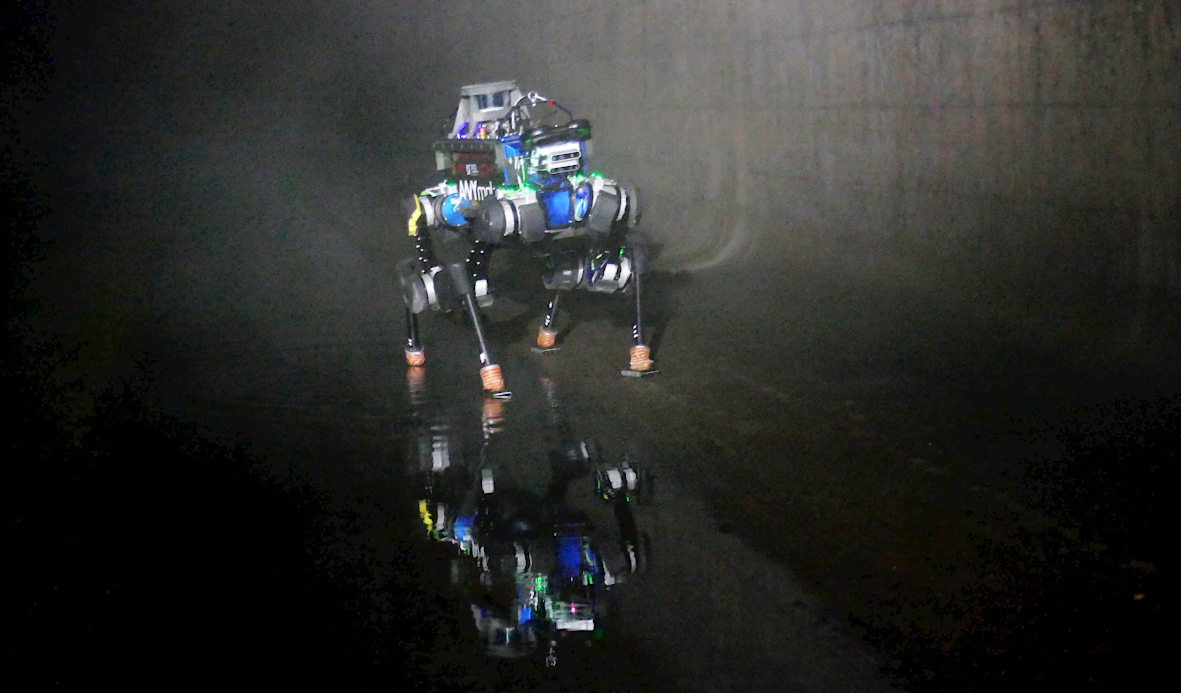}
 \caption{An ANYmal robot (\cite{hutter2016iros}) in a sewer with two feet in a slippery, wet depression and two feet in a dry, elevated area. With prior information about terrain type and geometry, it is possible for the robot to localized in the world using only touch. This would be extremely useful in dark and foggy environment (Image courtesy of RSL/ETH).}
\label{fig:sewer}
\end{figure}

Blind quadrupedal locomotion has achieved impressive levels of reactive robustness without requiring vision sensors (\cite{Focchi2020,hutter2020science}). However, without the ability to also \textit{localize} proprioceptively, a robot would still be incapable of executing goals and completing missions or inspections.

Availability of a prior map is a reasonable assumption for inspection tasks in engineered environments. In our previous work (\cite{buchanan2020haptic}), we demonstrated  that proprioceptive localization could be achieved online when prior geometric knowledge about the environment is available. However, the terrain morphology must include sufficiently informative features (\eg irregular steps) such that a robot can use them as landmarks for localization. 

To overcome this limitation, in this paper we propose a novel localization system that uses Sequential Monte Carlo at its core to fuse geometric as well as tactile semantic information from a terrain classifier to effectively and reliably localize the ANYmal B300 quadruped robot (\cite{hutter2016iros}) using limited sensing data.

\subsection{Contributions}
The contributions of this paper can be summarized as follows:
\begin{itemize}
 \item A novel proprioceptive legged robot localization system that fuses terrain geometry and semantics. The localization is performed in all the 6-DoF of the robot, instead of the x, y and yaw dimensions as in previous works. To the best of the author's knowledge, this is also the first such system that performs localization using semantics when completely blind. 
\item A terrain classification method employing signal masking in the 1D convolutional modules, making it possible to process variable length signals from footsteps without the need to truncate or pad them.
\item Extensive testing on an ANYmal B300 quadruped robot in three experiments including different geometries (both 2.5D and 3D) and terrain types, for a total duration of \SI{2.5}{\hour} and more than \SI{1.2}{\kilo\meter} of traveled distance. In 2.5D experiments localization error is kept down to \SI{10}{\centi\meter} on feature rich terrain and on average below \SI{20}{\centi\meter} on all terrain while exploiting terrain semantics. The localization is used online to effectively guide the robot towards its planned goals. The algorithm was also demonstrated to work with 3D maps, localizing by probing vertical surfaces.
\end{itemize}


The remainder of this document is structured as follows:
Section~\ref{sec:related} summarizes relevant research in the fields of terrain classification, legged haptic localization and the related problem of in-hand tactile localization; Section~\ref{sec:problem} defines the mathematical background of the legged
haptic localization problem; Section~\ref{sec:proposed} describes our proposed haptic localization algorithm; Section~\ref{sec:implementation} describes the implementation details to deploy our algorithm on a quadruped platform; Section~\ref{sec:experimental} presents the experimental results collected using the ANYmal robot; Section~\ref{sec:discussion} provides an interpretation of the results and discusses the limitations of the approach; finally, Section~\ref{sec:conclusion} concludes the paper with final remarks.

\section{Related Works}
\label{sec:related}

Pioneering work which exploits a robot's legs, not just for locomotion, but also to infer terrain information such as friction, stiffness and geometry has been presented by \cite{krotkov1990active}. This idea has recently been revisited to perform terrain vibration analysis or employ terrain classification to improve locomotion parameter selection. Since we are interested in using terrain classification for localization, we cover the most relevant works applied to legged robots in Section~\ref{sec:tactile-classification}. Works on proprioceptive localization in manipulation and legged robots are described in Sections \ref{sec:tactile-localization} and \ref{sec:haptic-localization}, respectively.

\subsection{Tactile Terrain Classification}
\label{sec:tactile-classification}
The first tactile terrain classification method for walking robots was presented by \cite{Hoepflinger2010clawar} and concerned experiments with a single leg detached from a robot's body. This seminal work utilized force measurements and motor currents to successfully distinguish between four terrain types. These results paved the way for the application of terrain classification methods on actual legged robots.

More recently, \cite{kolvenbach19} also used a single leg on a real, standing ANYmal robot to differentiate between four different types of soil by probing. Two types of feet (point and planar) were used to collect force, torque and IMU measurements that were processed by an SVM classifier.
The system showed that the tactile information can be used to differentiate between visually similar soils but the probing action is not particularly practical as it controls each part of the legged robot and prevents the system from performing other tasks at the same time.


A terrain classification system that could operate while the robot walks was presented by \cite{Wellhausen19}.
At the beginning of operation, their legged robot is trained to assign a terrain negotiation cost based on force/torque sensors.
Once operating, their system assigns a terrain negotiation cost from images based on previous feet-to-image correspondences and terrain classification based on proprioceptive sensors.
The ability to predict the terrain negotiation based on images is then used to plan robot's movement and avoid high cost terrains.

However, in complete darkness, which is our intended domain, vision-based sensors are of limited use for terrain classification, so we focus more on purely proprioceptive sensing.
In our previous work (\cite{bednarek19}), we showed how more complex and data-intensive deep learning models can be used to increase  terrain classification accuracy. This system achieved an accuracy of almost $98\%$ with six terrain classes and measurements taken from force-torque sensors during a statically stable walk. However, this approach was limited to fixed-length input signals, and thus did not generalize well to irregular walking patterns, different speeds, or walking on slopes, as they would produce variable length signals. The dependency on fixed-length inputs requires truncation and consequently information loss. In this work, we develop a novel masking mechanism in the convolutional layer, so as to process signals of variable lengths.

More recently, \etalcite{Lee}{hutter2020science} showed a completely different approach to terrain classification for locomotion. Their idea is to use an end-to-end deep learning controller based on proprioceptive signals to adapt the gait to rough terrains. Such a system does not explicitly perform tactile terrain classification, but an internal representation of the terrain type is implicitly stored inside the memory of the network.
In our work, we opt for a modular approach that is not based on end-to-end deep learning. We develop a deep learning module that explicitly returns a terrain class, which can then be used by the localization
estimator. 

\subsection{Tactile Localization in Manipulation}
\label{sec:tactile-localization}
Tactile localization involves the estimation of the 6-DoF pose of an object (of known shape) in the robot's base frame by means of kinematics of the robot's fingers and its tactile sensors. The pose of the object (in robot base coordinates) is inferred by matching the set of all positions of the sensed contacts (also in base coordinates) with the 3D model of the object being grasped.

Since the object can have any shape, the probability distribution of its pose given tactile measurements can be multimodal. 
For this reason, tactile localization has typically been addressed using Sequential Monte Carlo (SMC) methods, a subfamily of which are called \emph{particle filters} (\cite{Fox2001}). In contrast to Gaussian-based methods, SMC can maintain a discrete approximation of an arbitrary distribution by generating many hypotheses (often called particles) from a known \emph{proposal distribution}. If the number of particles is large enough and the proposal distribution can cover the state space well (i.e., it captures the areas of high density of the unknown underlying distribution), the target distribution will be well approximated by the set of state particles and their associated importance weights. SMC is, however, sensitive to the dimension of the state space, which should be low enough to
avoid combinatorial explosions or particle depletion. State-of-the-art methods aim to reduce this dimensionality and also to sample the state space in an efficient manner. In the SLAM community, FastSLAM achieved this by careful design of the proposal distribution and adaptive importance resampling to avoid particle depletion (\cite{Montemerlo2007}).

\cite{vezzani2017tro} proposed an algorithm for tactile localization using the Unscented
Particle Filter (UPF) and tested it on the iCub robot (equipped with contact sensors at the fingertips) to localize four different objects in the robot's reference frame. The algorithm is recursive and can process data in real-time as new measurements become available. The object and the robot's base are assumed to be static, allowing the pose to be estimated as a fixed parameter. This assumption allows the use of a window of
past measurements to better address the sparsity of the measurements, which consist of a series of single finger touches. For legged haptic localization, the assumption of both a static robot and terrain does not always hold and more general methods are required.

\cite{chalon2013online} proposed a particle filtering method for online in-hand object localization that tracks the pose of an object while it is being manipulated by a fixed base arm. The estimated pose is subsequently used to improve the performance of pick and place tasks (e.g., by placing a bottle in a upright position). The particles representing the object's pose are initialized by a Gaussian distribution around the true initial pose of the object, acquired by a vision system (used only at start). The particle weights are updated by penalizing finger/object
co-penetration and the distance between the object and the fingertip that detected the contact. Finally, their system selects the particle with the highest weight as the output estimate. 

To adapt the tactile localization problem to legged robots, one can instead imagine the robot as a hand trying to ``grasp'' the ground. The objective of legged haptic localization is estimating the pose of the robot relative to a fixed object (in this case the terrain) by means of its ``fingers'' (in this case the robot's feet), instead of the pose of the object being grasped relative to the robot. This problem is generally harder, because a legged robot cannot envelop the entire terrain at once as a hand can do with an object. This implies more uncertainty, as the robot needs to constantly move to collect enough informative samples.

\subsection{Haptic Localization of Legged Robots}
\label{sec:haptic-localization}
The first example of haptic localization applied to legged robots is from \etalcite{Chitta}{chitta2007icra}. In their work, they presented a proprioceptive localization algorithm based on a particle filter for the LittleDog, a small electric quadruped. The robot was commanded to perform a statically stable gait over a known irregular terrain course, using a motion capture system to feed the controller. While walking, the algorithm approximated the probability distribution of the base state with a set of particles. The state was reduced from six to three dimensions: linear position on the $xy$-plane and yaw. Each particle was sampled from the uncertainty of the odometry, while the weight of a particle was determined by the L2 norm of the height error between the map and the contact location of the feet. The algorithm was run offline on eight
logged trials of \SI{50}{\second} each.

\cite{schwendnerJfr} demonstrated haptic localization on a wheeled robot with protruding spikes, which act as a set of passive and rigid legs. The spikes detected contacts with the ground, which were compared to a prior 2.5D elevation map. The unique design of their robot enabled multiple contact measurements at a high rate for each wheel. They used the many contacts to perform plane fitting against the prior map and improve localization over larger, flatter terrain. They also performed terrain classification but with the use of a camera, which we do not require in our proposed work. They demonstrated an average position error \SI{39}{\centi\meter} in five experiments, each approximately \SI{100}{\meter}.

In \cite{buchanan2020haptic}, we presented an SMC method that estimates the past trajectory at every step instead of estimating just the most recent pose. This enabled the estimation of poses which are globally consistent. Furthermore, the localization was performed for the full 6-DoF of the robot, instead of just the $x$, $y$ and yaw dimensions as in \cite{chitta2007icra} and \cite{schwendnerJfr}. The localization system was experimentally demonstrated online and onboard an ANYmal robot and used in a closed loop navigation system to successfully execute a planned path, with a localization accuracy of \SI{10}{\centi\meter} on feature rich terrain. When walking on flat areas, the localization uncertainty increased due to the lack of constraints on the $xy$-plane. This behavior was expected, but undesirable. We are therefore motivated to use terrain
classification techniques described in Section \ref{sec:tactile-classification} to incorporate more information. 


\section{Problem Statement}
\label{sec:problem}
Let $\boldsymbol{x}_k = \left[ \mathbf{p}_k, \boldsymbol{\theta}_k\right]$ be
a robot's pose at time $k$, composed of the base
position in the world frame $\mathbf{p} \in \mathbb{R}^3$ and orientation
$\boldsymbol{\theta} \in
SO(3)$. With a slight abuse of notation, we will use the same symbol for its
homogeneous matrix form $\boldsymbol{x}_k \in SE(3)$.

\subsection{Quadruped State Definition}
We assume that for each timestep $k$, an estimate of the robot pose
$\tilde{\boldsymbol{x}}_k$ and its covariance $\Sigma_k \in
\mathbb{R}^{6\times6}$ are available from an inertial-legged odometric
estimator, such as \cite{bloesch2017ral, finkiros2020, hartley}. The uncertainties
for the rotation manifold are maintained in the Lie tangent space, as in
\cite{forster2017tro}. We also assume that the location of the robot's end
effectors in
the base frame 
$\mathcal{D}_k
= (\mathbf{d}_\text{LF},\;\mathbf{d}_\text{RF},\;\mathbf{d}_\text{LH } ,
\;\mathbf { d
}
_\text{RH}) \in \mathbb{R}^{3 \times 4}$ are known from forward
kinematics and their binary contact states $\mathcal{K}_k = (\kappa_\text{LF},\kappa_\text{RF},\kappa_\text{LH},\kappa_\text{RH}) \in
\mathbb{B}^4$ from inverse dynamics. Finally, where available, a foot sensor measures the force acting on each foot $\mathcal{F}_k
= (\mathbf{f}_\text{LF},\;\mathbf{f}_\text{RF},\;\mathbf{f}_\text{LH},
\;\mathbf { f
}
_\text{RH}) \in \mathbb{R}^{3 \times 4}$.


For simplicity,
we neglect any uncertainties due to inaccuracy in joint encoder readings or limb
flexibility. Therefore,
the propagation of the uncertainty from the base to the end effectors is
straightforward to compute. For brevity, the union of the aforementioned states
(pose, contacts, and forces)  at time $k$ will be referred as the
\emph{quadruped state}
$\mathcal{Q}_k = \{
\tilde{\mathbf{p}}_k, \tilde{\boldsymbol{\theta}}_k, \Sigma_k,
\mathcal{D}_k,\;
\mathcal{K}_k,\;\mathcal{F}_k\}$.

\subsection{Prior Map}

\begin{figure}
\centering
 \includegraphics[width=0.9\columnwidth]{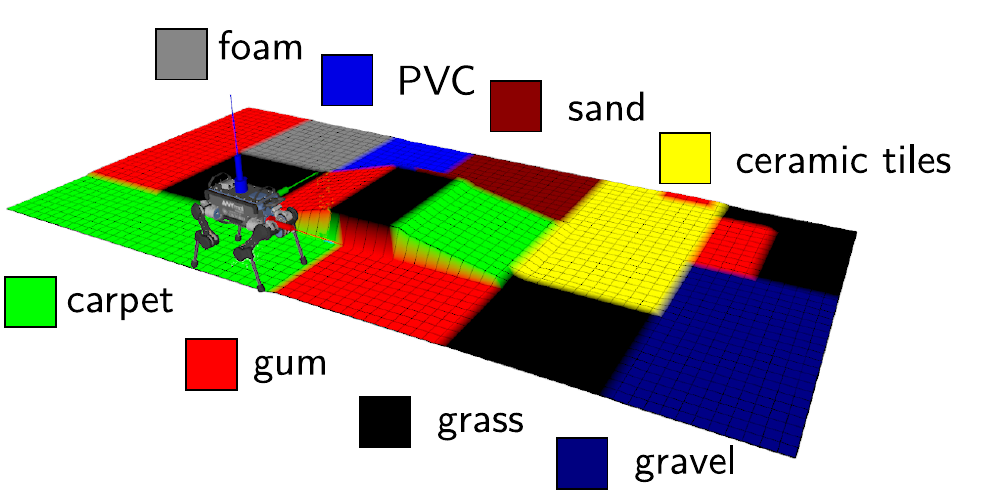}
 \caption{Terrain map used in Experiment 3, showing terrain categories.}
 \label{fig:augmented}
\end{figure}

Our approach can localize against 2.5D terrain elevation maps as well as full 3D maps. Terrain classification is meant to be carried out while the robot is walking, with no dedicated probing actions, therefore 2.5D maps are augmented with a terrain class category for each cell. This enables our method to overcome the degeneracy caused by featureless geometries (\eg flat grounds, which are uninformative about the robot position on the $xy$-plane). Point clouds are used for 3D maps and only contain geometric information. To distinguish between the three types of map, we will refer to $\mathcal{M}$ when 2.5D only, $\mathcal{M}_3$ when 3D, and $\mathcal{M}_{c}$ when 2.5D augmented with class information, respectively. An example of an $\mathcal{M}_{c}$ map colorized by terrain class is shown in Figure \ref{fig:augmented}.

\subsection{Estimation Objective}
Our goal is to use a sequence of quadruped states and their corresponding
uncertainties
$\Sigma_k$ to estimate the most likely sequence of robot states up to time $k$
as:
\begin{equation}
\mathcal{X}^{*}_{k} =
[\boldsymbol{x}^{*}_0,\;\boldsymbol{x}^{*}_1,\;\dots,\;\boldsymbol{x}^{*}_k]
\end{equation}
such that the likelihood of the contact points to be on
the map is maximized.


\section{Proposed Method}
\label{sec:proposed}
To perform localization, we sample a predefined number of particles at
regular intervals from the pose distribution provided by the odometry (as
described in Section~\ref{sec:sampling}) and we compute the likelihoods of
the measurements by comparing each particle to the prior
map, so as to update the weights of the particle estimator (see
Section~\ref{sec:measurement}).

%

\subsection{Locomotion Control and Sampling Strategy}
\label{sec:sampling}

Blind locomotion requires cautious footstep placement, hence we opt for a statically stable gait, which guarantees stability at all times even when the motion is stopped mid flight phase.
Since only one leg can be moved at the time, as soon as the swing leg touches the ground the robot enters into a four-point support phase; at this time, the quadruped state estimate $\mathcal{Q}_k$ and the estimated terrain class $\tilde{c}$ for a given foot position are collected.
Then, a new set of particles is sampled in a manner similar to \cite{chitta2007icra}:

\begin{equation}
 p(\boldsymbol{x}_k | \boldsymbol{x}_{k-1}^i) =
\mathcal{N}(\boldsymbol{x}_k, \Delta\tilde{\boldsymbol{x}}_{k}
\boldsymbol{x}_{k-1}^i, \Sigma_k)
\end{equation}
where $\Delta\tilde{\boldsymbol{x}}_{k} =
\tilde{\boldsymbol{x}}_{k-1}^{-1}\tilde{\boldsymbol{x}}_{k}$ is the pose
increment measured by the onboard state estimator at times $k-1$ and $k$.

At time $k$, the new particles $\boldsymbol{x}_k^i$ are thus sampled from a normal distribution centered at the pose estimated from the odometry with its corresponding covariance. Since roll and pitch angles are observable from inertial-legged estimators, their estimates have low uncertainty. In practice, this allows us to reduce the number of necessary particles along these two dimensions, mitigating issues related to high-dimensional sampling in 6-DoF.

\subsection{Measurement Likelihood Model for 2.5D Data}
\label{sec:measurement}
\begin{figure}
 \centering
\includegraphics[width=\columnwidth]{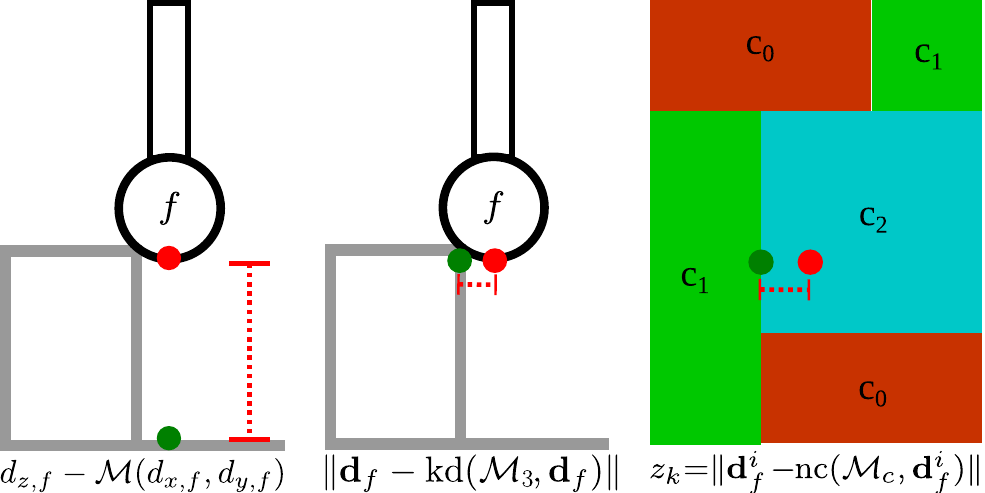}
 \caption{Comparison between contact measurements for 2.5D (left), 3D (middle)
and terrain class (right) map representations. The red dots indicate the
contact point as sensed by the robot, while the green dots indicate the
corresponding location returned by the map. The red line shows the magnitude of
the measurement. In the terrain classification case, the view is from a top down perspective. The robot's foot sensed a contact in the $c_2$ region but a
classification of $c_1$ was detected. The nearest $c_1$ point was returned by
the map.}
 \label{fig:error}
\end{figure}

We use the same measurement model for 2.5D measurements as in our previous
work (\cite{buchanan2020haptic}). The measurement likelihood is modeled as a
univariate Gaussian centered at the local elevation of each cell. We use a
manually selected variance $\sigma_z$ of \SI{1}{\centi\meter}. Given a particle
state $\boldsymbol{x}^i_k$, the estimated position of a contact in world
coordinates for an individual foot $f$ is defined as the concatenation of the
estimated robot base pose and the location of the end effector, in base
coordinates:
\begin{equation}
\mathbf{d}^i_f = (d^i_{xf}, d^i_{yf}, d^i_{zf},) =
\boldsymbol{x}^i_k \mathbf{d}_f
\end{equation}
Thus, the measurements and their relative likelihood functions for the
$i$-th particle and a specific foot $f$ are (Figure \ref{fig:error},
left):
\begin{align}
 z_k &= d^i_{zf} - \mathcal{M}(d^i_{xf},d^i_{yf})\\
p(z_k | \boldsymbol{x}^i_k) &=
\mathcal{N}(z_k,0,\sigma_z)
\label{eq:2dlikelihood}
\end{align}
where: $d^i_{zf}$ is the vertical component of the estimated contact
point location in world coordinates of foot $f$,  according to the $i$-th
particle; $\mathcal{M}(d^i_{x,f},d^i_{y,f})$ is the
corresponding map elevation at the  $xy$ coordinates of
$\mathbf{d}_f^i$.

\subsection{Measurement Likelihood Model for 3D Data}
\label{sec:measurement3d}
Our method can incorporate contact events from 3D probing. This is useful for areas where the floor does not provide enough information to localize. In this case, the robot can probe walls and 3D objects with its feet.
To better model this situation, we represent the prior map $\mathcal{M}_3 \in
\Real^{3 \times N}$ by a 3D point cloud with $N$ points.
The likelihood of a particular contact point is computed using the Euclidean
distance
between the foot and the nearest point in the map.
This likelihood is again modeled as a zero-mean Gaussian
evaluated at the Euclidean distance between the estimated contact point
$\mathbf{d}^i_{f}$ and its nearest neighbor on the map, with variance
$\sigma_z$:
\begin{align}
 z_k &= \lVert\mathbf{d}^i_{f}-\text{kd}(\mathcal{M}_3,\mathbf{d}^i_{f})\rVert
\\
p(z_k | \boldsymbol{x}^i_k) &=
\mathcal{N}(z_k,0,\sigma_z)
\label{eq:3dlikelihood}
\end{align}
where $\text{kd}(\mathcal{M}_3,\mathbf{d}^i_{f})$ is the function that returns
the nearest neighbor of $\mathbf{d}^i_{f}$ on the map $\mathcal{M}_3$,
computed from its k-d tree (Figure \ref{fig:error}, middle).

\subsection{Terrain classification}

We define the haptic terrain classification function $f\colon \calS \mapsto \calC$, as the function that associates an element from the signal domain $\calS$ to an integer from the class counter-domain $\calC$. The set $\calS: \{s: s \in \Real^{l(s) \times 6}\}$, where $l(s)$ is a length of the signal $s$, includes sequences of variable length signals such as the forces and torques (6 values in total) generated by a robot's foot touchdown event while taking a step. The set $\calC$ is defined as the integers from $0$ to $n-1$, where $n$ is the total number of terrain classes that the robot is expected to be walking on (in our case, $n = 8$).

Having defined the problem as such, we introduce the following:
\begin{itemize}
	\item a classification method $f^\prime\colon \calS \mapsto \calC$,
which approximates the function $f$. As an implementation of $f^\prime$
we used a neural network;
	\item a dataset consisting of a list of pairs $d\colon [(s, c)]$, where $s \in \calS$, $c \in
\calC$. Such a dataset was divided into two subsets, training and validation, with
a ratio of 80:20;
	\item a training process formulated as an approximation of the function $f$
using the function $f^\prime$ by the minimization of cross-entropy between the
probability distributions generated by these functions.
\end{itemize}

\begin{figure*}[t]
\centering
 \includegraphics[width=0.7\textwidth]{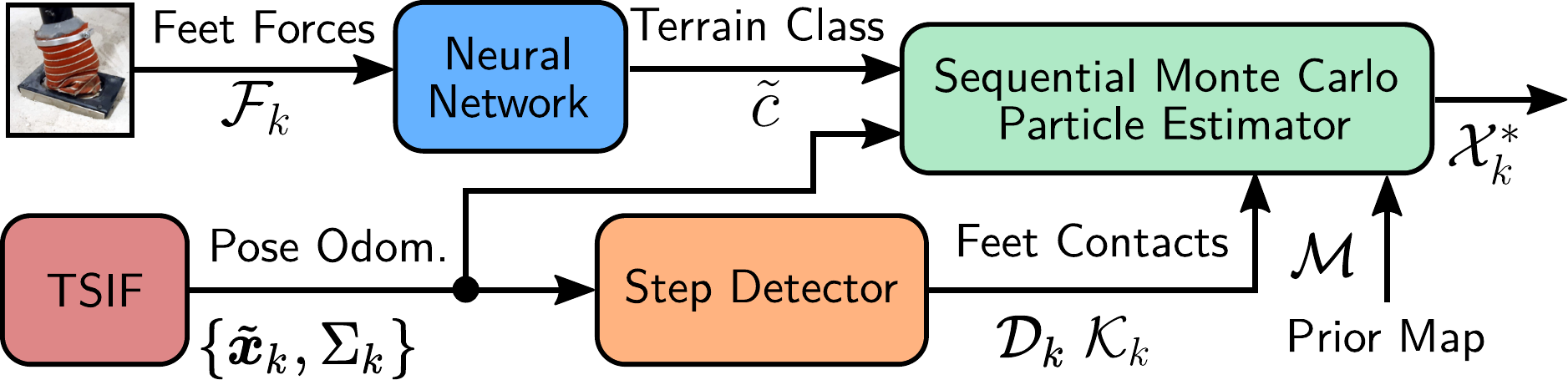}
 \caption{Information flow diagram of the overall system.}
 \label{fig:flor-diagram}
\end{figure*}

\subsection{Terrain Class Measurement Likelihood Model}
\label{sec:terrainClass}
To localize with the support of terrain classes, we represent the prior map as a 2-dimensional grid whose cells are associated with a terrain class. An example is provided by Figure \ref{fig:augmented}, where each cell is colorized according to its class.

Measurements are represented as a piece-wise cost function. In the case where the
estimated terrain class $\tilde{c}$ for a given foot position $\mathbf{d}_f^i$
does match the class $c$, at that location in the prior map $c =
\mathcal{M}_{c}(d^i_{xf},d^i_{yf})$ the probability is a constant value. This constant value is the maximum value of a zero-mean univariate Gaussian with manually selected variance $\sigma_c$ of \SI{5}{cm} (which is the width of the foot). 

If the estimated class does not agree with the expected class in the map, the same Gaussian distribution is used to model the likelihood, where the input $z_k^{c}$ is the distance to the closest point in the map with the expected class. This function is shown as
\begin{align}
p(z_k | \boldsymbol{x}^i_k)  = \begin{cases}
\frac{1}{\sigma_c\sqrt{2\pi}} & \tilde{c} = c\\
\mathcal{N}(z_k,0,\sigma) & \tilde{c} \neq c,
\end{cases}
\label{eq:terrainlikelihood}
\end{align}
where
\begin{align}
z_k^{c} &= \lVert\mathbf{d}^i_{xyf}-\mathrm{nc}(\mathcal{M}_{c},\mathbf{d}^i_{xyf})\rVert_2.
\label{eq:terrainlikelihood2}
\end{align}
The function $\mathrm{nc}(\mathcal{M}_{c},\mathbf{d}^i_{xyf})$ returns the nearest
2D point in the map $\mathcal{M}_c$ to the 2D foot position $\mathbf{d}^i_{xyf}$, which has the estimated class $\tilde{c}$. This last case is shown in Figure \ref{fig:error}, right.

We assume elevation and terrain class measurements ($z_k, z_k^c$) are conditionally independent. Therefore, their joint probability can be computed as
\begin{align}
p(z_k, z_k^c | \boldsymbol{x}^i_k) = p(z_k | \boldsymbol{x}^i_k) p(z_k^c | \boldsymbol{x}^i_k).
\label{eq:conditional-independance}
\end{align}

\section{Implementation}
\label{sec:implementation}
The block diagram of our system is shown in Figure \ref{fig:flor-diagram}. The internal estimator on the ANYmal robot, TSIF (\cite{bloesch2017ral}), provides the odometry for the particle estimator, while the neural network estimates the class. This information is compared against the prior map to provide an estimate of the robot's trajectory, $\mathcal{X}^*_k$. 

Pseudocode for the particle estimator (green block in Figure \ref{fig:flor-diagram}) is listed in Algorithm~\ref{alg:pf}.
At time $k$, the estimates of the terrain class $\tilde{c}$ and the robot pose $\tilde{\boldsymbol{x}_k}$  are collected. The pose estimate is used to compute the relative motion
$\Delta\tilde{\boldsymbol{x}}_{k-1:k}$, propagate forward the
state
of each particle $\boldsymbol{x}^i_k$, and draw a sample from the distribution
centered in $\Delta\tilde{\boldsymbol{x}}_{k}
\boldsymbol{x}_{k-1}^i$ with covariance $\Sigma_k =
(\sigma_{x,k},\sigma_{y,k},\sigma_{z,k})$.

The weight of a particle $w^i$ is then updated by multiplying it by the
likelihood that each foot is in contact with the map and the terrain class. In
our implementation, we
modify the likelihood functions from Equation \ref{eq:2dlikelihood} as:
\begin{equation}
p(z_k | \boldsymbol{x}^i_k) = \min(\rho,\mathcal{N}(z_k,0,\sigma_z))
\end{equation}
where $\rho$ is a minimum weight threshold, so that outlier contact measurements do not immediately lead to degeneracy. 

Resampling is triggered when the variance of the weights rises
above a certain threshold.  This is necessary to avoid dispersion of the particle set across the state space, with many particles with low weight. By triggering this process when the variance of the weights increases, the particles can first track the dominant modes of the underlying distribution.

\begin{algorithm}[!t]
$\boldsymbol{x}_0^i \sim \mathcal{N}(\boldsymbol{x}_0,
\tilde{\boldsymbol{x}}_{0}, \Sigma_0)\quad \forall i \in N$

\ForEach{four-support phase $k$}{
	$\Delta\tilde{\boldsymbol{x}}_{k} \leftarrow%
	\tilde{\boldsymbol{x}}_{k-1}^{-1}\tilde{\boldsymbol{x}}_{k}$

	\ForEach{particle $i \in N$}{
		$\boldsymbol{x}^i_k \sim \mathcal{N}(\boldsymbol{x}_k,
		\Delta\tilde{\boldsymbol{x}}_{k}
		\boldsymbol{x}_{k-1}^i, \Sigma_k)$
		
		$w^i_k \leftarrow w^i_{k-1}$
		
		\ForEach{foot $f$} {
		\eIf{3D}{
		    $z_k \leftarrow \lVert\text{kd}(\mathcal{M}_3,\mathbf{d}^i_{f}) -%
\mathbf{d}^i_{f}\rVert$

            $w^i_k \leftarrow w^i_k p(z_k | \boldsymbol{x}^i_k)$
		}{
		  $z_k \leftarrow d^i_{zf} - \mathcal{M}(d^i_{xf},d^i_{yf})$
		  
		  $z_k^{c} \leftarrow \lVert\mathbf{d}^i_{xyf}-\mathrm{nc}(\mathcal{M}_c,\mathbf{d}^i_{xyf})\rVert$
		  
		  $w^i_k \leftarrow w^i_k p(z_k, z_k^{c} | \boldsymbol{x}^i_k)$
		}
	    }
	$ x_k^* \leftarrow \text{WeightedMean}(x_k^0 \dots x_k^N, w_k^0,\dots
w_k^N)$

	$\mathcal{X}_k^* \leftarrow [\boldsymbol{x}_0^j,\dots, \boldsymbol{x}_k^j]$
	
	\If{$\text{Var}(w_k^i) > \text{threshold}$}
	{$\mathrm{resample}(x_k^i)$}

}
}
\caption{Haptic Sequential Monte Carlo Localization}
\label{alg:pf}
\end{algorithm}

\subsection{Particle Statistics}
\label{sec:statistics}

The pose estimate for the $k$-th iteration,
$\boldsymbol{x}^*_k$, is computed from the weighted mean of all the
particle poses. However, as we showed in our previous work, the particle
distribution is often multi-modal. This motivated us to selectively update different dimensions of the robot pose. If the variance of the particle positions in the $x$ and $y$ axes are low (\ie $\ll \sigma_{x,k},\sigma_{y,k}$), we assume a well defined estimate and
update the robot's full pose. However, if they are high, we update only the $z$ component of the robot's location, which is always low as the robot keeps contact with the ground. 

In practice, with the terrain classification we found the
terrain course was sufficiently detailed to keep particle position variance in the $xy$-plane low, therefore $z$-only updates were rare. The threshold we used was a standard deviation of \SI{10}{\centi\meter}, which corresponds to the typical uncertainty in our experiments. 

To avoid particle degeneracy, importance sampling can be done in areas with higher likelihood. For example, if a grass terrain is detected, some particles could be injected in every grass terrain in the map. Further investigation on the benefits of importance sampling are left for future work.

\subsection{Terrain Classifier Network}

\begin{figure}
 \centering
 \includegraphics[width=\columnwidth]{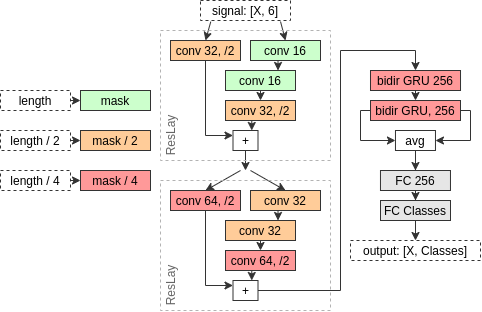}
 \caption{Neural Network Structure used for Terrain Classification on ANYmal. The main blocks of neural architecture are convolutional (conv), recursive (GRU bidir), and fully connected (FC) blocks. The masking mechanism used in variable-length signal processing blocks takes appropriate masks, marked in green, orange, and red. These masks correspond to the individual signal lengths, taking into account the initial length and the reduction of size by convolution with stride equal to 2.}
 \label{fig:network}
\end{figure}

The neural network architecture used for the terrain classification module is depicted in Figure \ref{fig:network} and consists of
three consecutive components: convolutional, recurrent, and predictive. Both
the convolutional and recurrent components process variable-length data.
Thus, for the convolution part to work properly, masking of the padded
values must be used to prevent these values from affecting the forward-pass
result.

In \cite{bednarek2019icra}, we have shown that both convolutional (if fixed-length signals are used) and recursive neural components can be successfully used to solve the terrain classification problem based on signals from force and torque sensors. Moreover, in  \cite{bednarek19} we introduced a model that uses both the convolutional and recursive components, achieving the best results in the classification of terrains. The problem of this architecture is the ability to process only fixed-length signals, so for the needs of our application we have adapted this model to processing of signals of variable length, adding the intermediate layer masking module.

The first component consists of two residual layers \newline (ResLay). The ResLay
used in our work is an adaptation of the layer used by \cite{He15Resnet} with
2D convolutions replaced by 1D and support masking. The recurrent component
uses two bidirectional layers (Bidir) with two GRU  (\cite{Cho14Gru}) cells in
each. The output of the recurrent component is an average of two resulting
hidden states of the last Bidir. The final output of the neural network is
produced by the predictive component, which takes the recurrent component's
output, and using two fully connected layers (FC) produces a probability
distribution from which classification is taken.

The consecutive layers of the model are presented in Figure
\ref{fig:network}.
Each convolution block executes the following operations: batch
normalization
(\cite{Ioffe15BatchNorm}), dropout (\cite{Srivastava14Dropout}),
and ELU activation function (\cite{Clevert2016FastAA}). All are
modified to support masking. We used kernel size of 5 in each
convolution layer. The output from each ResLay block is two times smaller thanks to the use of stride in convolutions (marked as $/2$). Dropout is also used in every Bidir and FC (which also use batch normalization).

The model uses a dropout rate of $0.3$ and a batch normalization momentum of $0.6$. The proposed neural network consists of $1,374,920$ trainable parameters.


\subsubsection{Training}
\label{sec:training}

The learning process was carried out using the k-fold cross-validation
methodology, with $k=5$. The AdamW optimizer from \cite{Loshchilov17AdamW} was used to minimize the loss function, with the following parameters:
\begin{itemize}
	\item learning rate: 5e-4 with exponential decay,
	\item weight decay: 1e-4 with cosine decay.
\end{itemize}

The learning process lasted for at least 1000 epochs, after which
the process continued until the result improved for the last 100
epochs. The size of each mini-batch was 256. 


%

\section{Experimental Results}
\label{sec:experimental}
We extensively evaluated the performance of our algorithm in three different experiments, each one targeting a different type of localization. These are described in more detail in Sections \ref{sec:experimental-2.5d}, \ref{sec:experiment-3d}, and \ref{sec:experiment-class}.

\subsection{Evaluation Protocol}
There are two different modalities in our algorithm: HL-G (Haptic Localization Geometric), which uses only geometric information, and HL-GC (Haptic Localization Geometric and Class), which uses both geometric and terrain class information.

HL-G was tested in Experiments 1 and 2 using 2.5D and 3D prior maps, respectively. HL-GC was tested in Experiment 3 with 2.5D maps augmented with terrain class information. We also compared the performance against the HL-G modality in Experiment 3.

\subsubsection{Evaluation Metrics}
We quantitatively evaluated localization performance by computing the the mean of the Absolute Translation Error (ATE) as described by \cite{benchmark}:
\begin{equation}
    \frac{1}{n}\sum_{i=1}^{n} \lVert\text{trans}\left(\mathbf{T}_i^{-1}\hat{\mathbf{T}}_i\right)\rVert
\end{equation}
where $\mathbf{T}_i$ and $\hat{\mathbf{T}}_i$ are the robot's ground truth and estimated poses, respectively.
In contrast to~\cite{benchmark}, we do not perform the alignment of trajectories as robot's ground truth and estimated poses are represented in the same coordinate system.

A qualitative evaluation was also performed for Experiment 1 and 2
by assessing the ability of the robot to reach its planned goals
or end effector targets while using the localization online. This
demonstrated the benefit of the localization when used in the loop
with the onboard motion planner.

\subsubsection{Ground Truth and Prior Map}
The ground truth trajectories were collected by motion capture systems at \SI{100}{\hertz}. Both the robot and the terrain courses were accurately tracked with \si{\milli\meter} accuracy via reflective markers installed on them.

At start of the experiment, the relative position of the robot within the map was measured using ground truth and used for initialization only. Thereafter, the pose of the robot was estimated using the particle filter. To account for initial errors, particles at the start were sampled from a Gaussian centered at the initial robot pose with a covariance of \SI{20}{\centi\meter}.


The prior maps were captured with survey grade laser scanners (Leica BLK-360 and SURPHASER 100HSX) which provide registered point clouds with sub-centimeter accuracy.


\begin{table}
	\vspace{4pt}
\centering
\begin{tabular}{ccccc}
\toprule
\multicolumn{5}{c}{\textbf{Mean Absolute Translation Error (ATE)  
[\si{\meter}]}} \\
\midrule
\textbf{Trial} & \textbf{Dist. [\si{\meter}]} & \textbf{Time [\si{\second}]} & \textbf{TSIF} & \textbf{HL-G} \\
\toprule
1 & 66.42 & 525 & 0.63 & \textbf{0.13} \\
\midrule
2 & 145.31 & 1097 & 2.57 & \textbf{0.40}  \\
\midrule
3 & 55.67 & 557 & 0.52   & \textbf{0.19} \\
\midrule
4 & 68.71 & 604 & 0.65 & \textbf{0.32}  \\
\midrule
5 & 172.65 & 1606 & 2.00 & \textbf{0.61}  \\
\bottomrule
\end{tabular}
 \caption{Experiment 1: Estimation performance. TSIF = Two-State Implicit Filter (\cite{bloesch2017ral}); HL-G = Haptic
Localization with Geometry only. }
\label{tab:rpe1}
\end{table}

\subsection{Experiment 1: 2.5D Terrain Course}
\label{sec:experimental-2.5d}
\begin{figure}
	\vspace{4pt}
	\centering
	\includegraphics[width=\columnwidth]{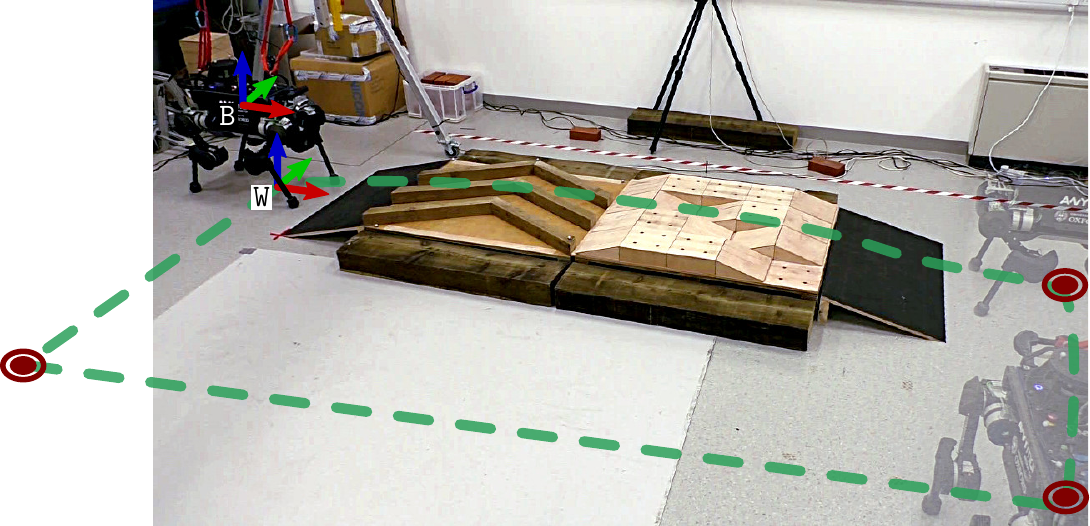}
	\caption[ANYmal haptic localization experiment]{Experiment 1: ANYmal haptic
		localization experiments. The robot traverses the terrain, turns 90 deg right and comes back to the initial position passing trough the flat area. The goals given to the planner are marked by the dark red disks, while the planned route is a dashed green line (one goal is out of the camera field of view). The world	frame $\World$ is fixed to the ground, while the base frame $\Base$ is rigidly	attached to the robot's chassis. The mutual pose between the robot and the
		terrain course is bootstrapped with the motion capture.}
	\label{fig:anymal-chevron}
\end{figure}

In this experiment, the robot was commanded to navigate between four walking  goals at the corners of a rectangle. One of the edges required crossing  a \SI{4.2}{\meter} terrain course composed of a
\SI{12}{\degree} ascending ramp, a \SI{13}{\centi\meter} high chevron pattern,
an asymmetric composition of uneven square blocks and a
\SI{12}{\degree} descending ramp (Figure \ref{fig:anymal-chevron}). After
crossing the wooden course, the robot returned to the starting
position across a portion of flat ground, which tests how the system behaves in feature-deprived conditions.

While blind reactive locomotion has been developed by a number of
groups including \cite{dicarlo2018iros} and \cite{Focchi2020}, unfortunately our blind controller was not sufficiently reliable to cross this terrain course so we resorted to use of the statically stable gait from \cite{fankhauser2018icra} which used a depth camera to aid footstep planning. However, the localization was performed without access to any camera information.

To demonstrate repeatability, we performed five trials of this experiment,
for a total distance traveled of more than \SI{0.5}{\kilo\meter} and \SI{1}{\hour}\SI{13}{\minute} duration. A summary of the experiments is presented in Table \ref{tab:rpe1}, where HL-G shows an overall improvement between \SIrange{50}{85}{\percent} in the Absolute Translation Error (ATE) compared to the onboard state estimator. ATE is \SI{33}{\centi\meters} on average, which reduces to \SI{10}{\centi\meters} when evaluating only the feature-rich portion of the experiments (\ie the terrain course traversal).

\begin{figure}
	\vspace{4pt}
 \centering
 \includegraphics[width=\columnwidth]{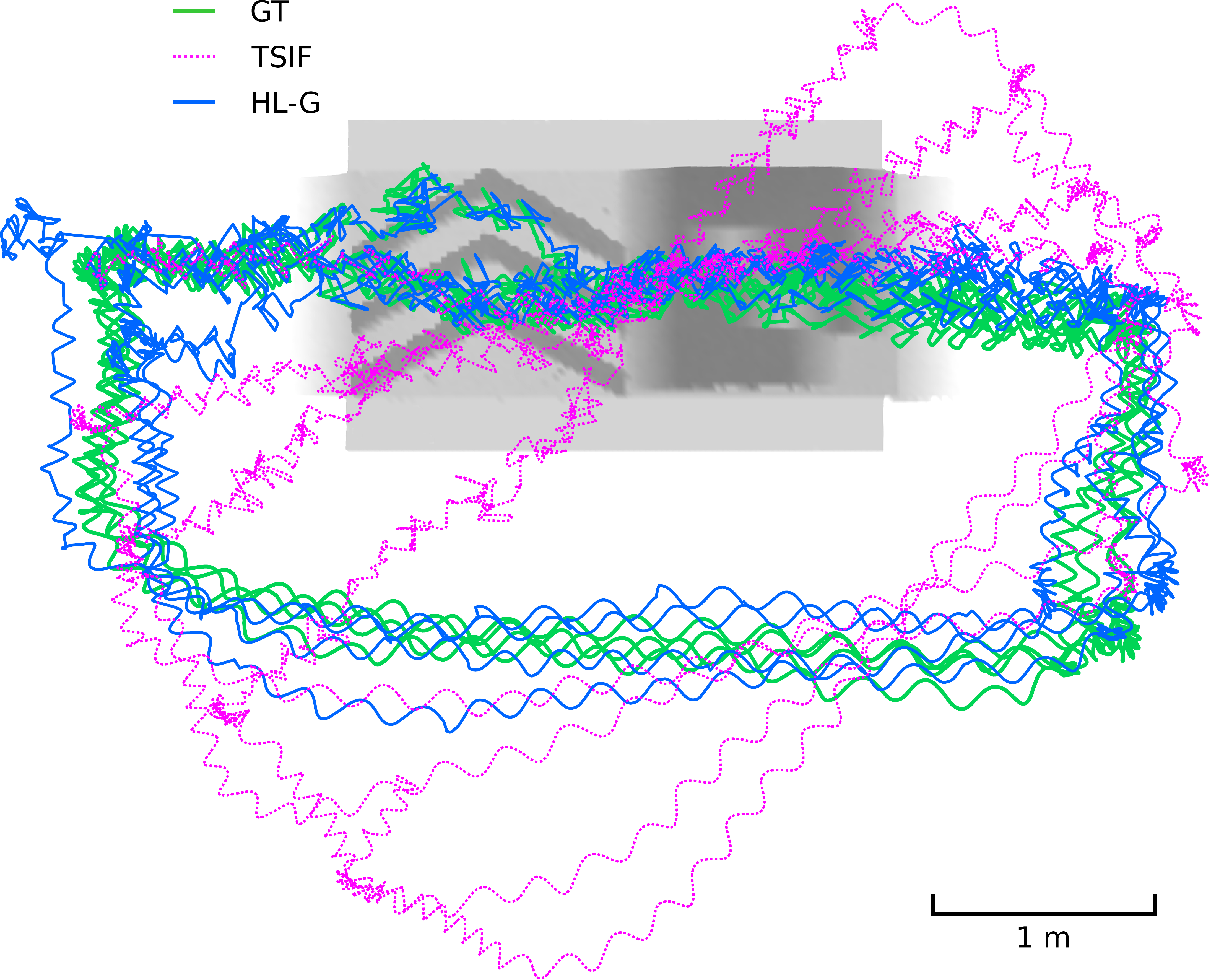}
 \caption{Experiment 1: Top view of the estimated trajectories from TSIF (dashed magenta),
haptic localization (blue) and ground truth (green) for Experiment 2.}
 \label{fig:top-view}
 \vspace{-10pt}
\end{figure}

For trials 1 and 2, the robot was manually operated to traverse the terrain course, completing two and four loops, respectively. In trials 3--5, the robot was commanded to follow the rectangular path autonomously. In these trials, the haptic localization algorithm was run online in closed-loop and effectively guided the robot towards the goals (Figure \ref{fig:top-view}). Using only the prior map and the contact events only, the robot stayed localized in all the runs and successfully tracked the planned goals. This can be seen in Figure~\ref{fig:top-view}, where the estimated trajectory (in blue) diverges from ground truth on the $xy$-plane when the robot is walking on the flat ground. This is due to growing uncertainty from lack of geometric information, however the covariance reduces significantly and the cluster mean re-aligns with the ground truth when the robot reaches the terrain course.

\begin{figure}
	\vspace{4pt}
 \centering
 \includegraphics[width=\columnwidth]{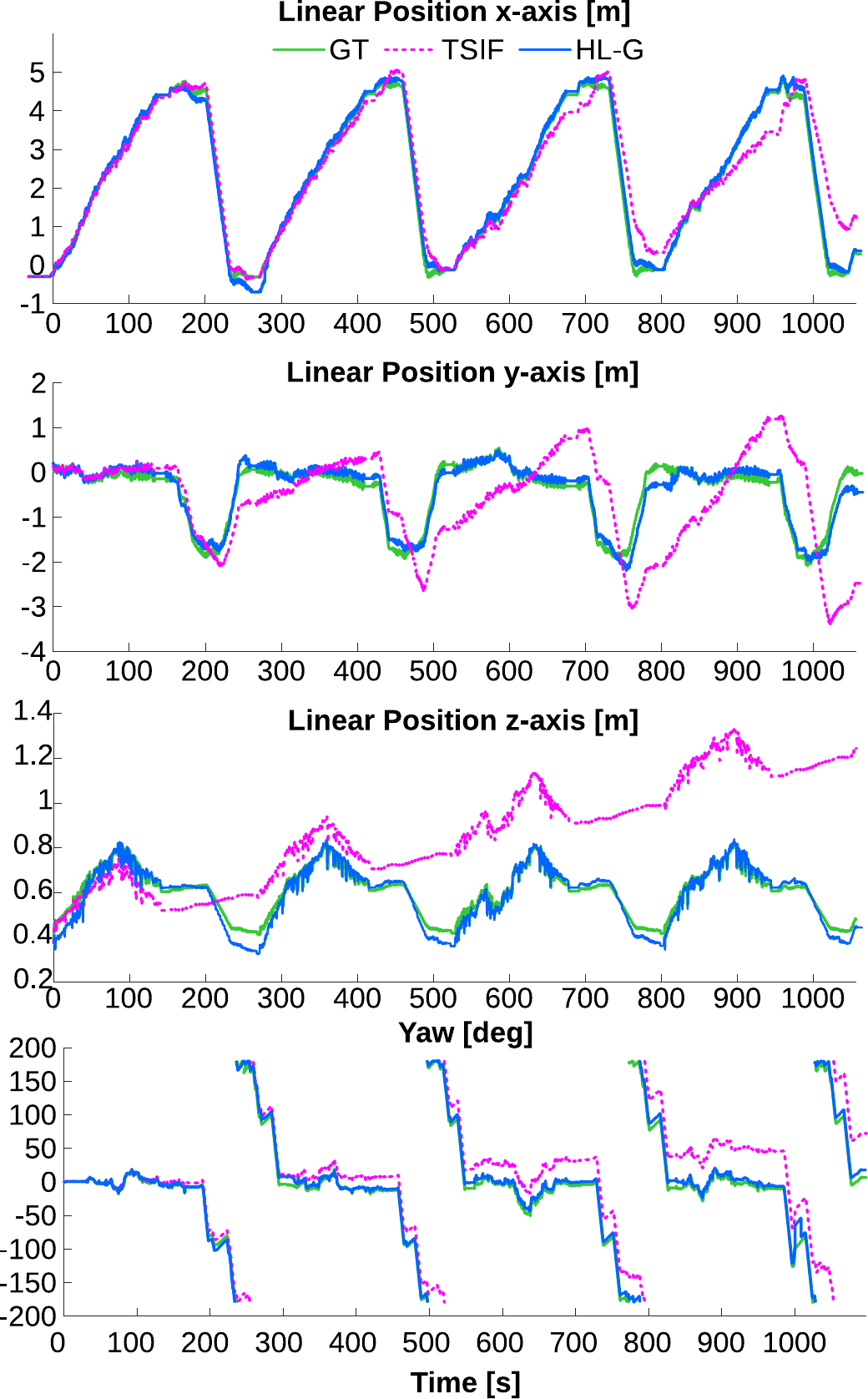}
 \caption{Experiment 1: Comparison between the estimated position from TSIF (dashed magenta)
and haptic localization (blue) against Ground Truth (green) for Experiment 2.
After \SI{200}{\second}, the estimation error in TSIF has drifted significantly upward and in yaw. In particular, the upward drift is noticeable in the third plot, where the values grow linearly. The drift is eliminated by the re-localization against
the prior map.}
 \label{fig:plots}
 \vspace{-10pt}
\end{figure}

Figure \ref{fig:plots} shows in detail the estimator performance for each of
the three linear dimensions and yaw. Since position and yaw are unobservable,
the drift on these states is unbounded. In particular, the error on the odometry filter (TSIF \cite{bloesch2017ral}, magenta dashed line) is dominated by upward drift (due to kinematics errors and impact nonlinearities, see third plot) and yaw drift (due to IMU gyro bias, see bottom plot). This drift is estimated and compensated for by the haptic localization (blue solid line), allowing accurate tracking of the ground truth (green solid line) in all dimensions. This can be noted particularly at the four peaks in the $z$-axis plot, where the estimated trajectory and ground truth overlap. These times coincide with the robot is at the top of the terrain course.

\subsection{Experiment 2: Online Haptic Exploration on Vertical Surfaces}
\label{sec:experiment-3d}

The second experiment involved a haptic wall following task with the robot starting in front of a wall but with an uncertain location. The particles were again initialized with \SI{20}{\centi\meter} position covariance. To test the capability to recover from an initial error, we applied a \SI{10}{cm} offset in both $x$ and $y$ from the robot's true position in the map. At start, the robot was
commanded to walk \SI{1}{\meter} to the right (negative $y$ direction) and press a button on the wall. To accomplish the task, the robot needed to ``feel its way'' by alternating probing motions with its right front foot and walking laterally to localize inside the room. The fixed number of probing motions was pre-scripted so with each step to the right, the robot probed both in front and to its right. The whole experiment was executed blindly with the static controller from \cite{fankhauser2018icra}.




\begin{figure}
	\vspace{5pt}
	\centering
	\includegraphics[width=0.240\columnwidth]{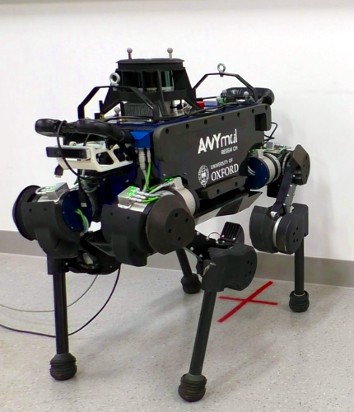}
	\includegraphics[width=0.240\columnwidth]{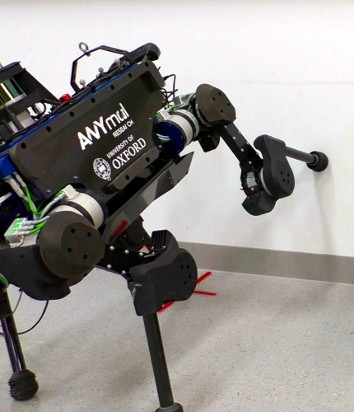}
	\includegraphics[width=0.240\columnwidth]{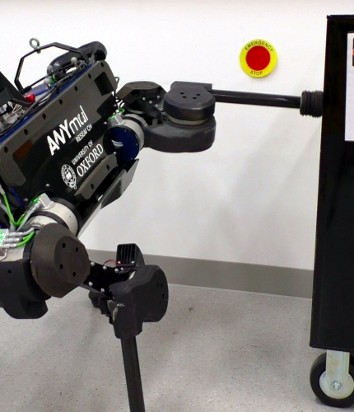}
	\includegraphics[width=0.240\columnwidth]{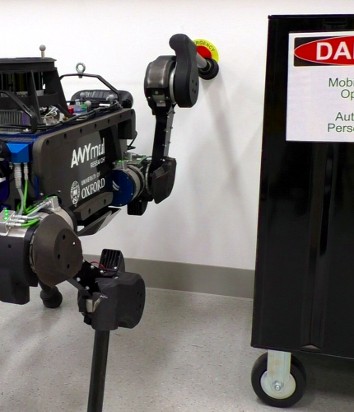}\\
	\vspace{2pt}
	\includegraphics[width=0.240\columnwidth]{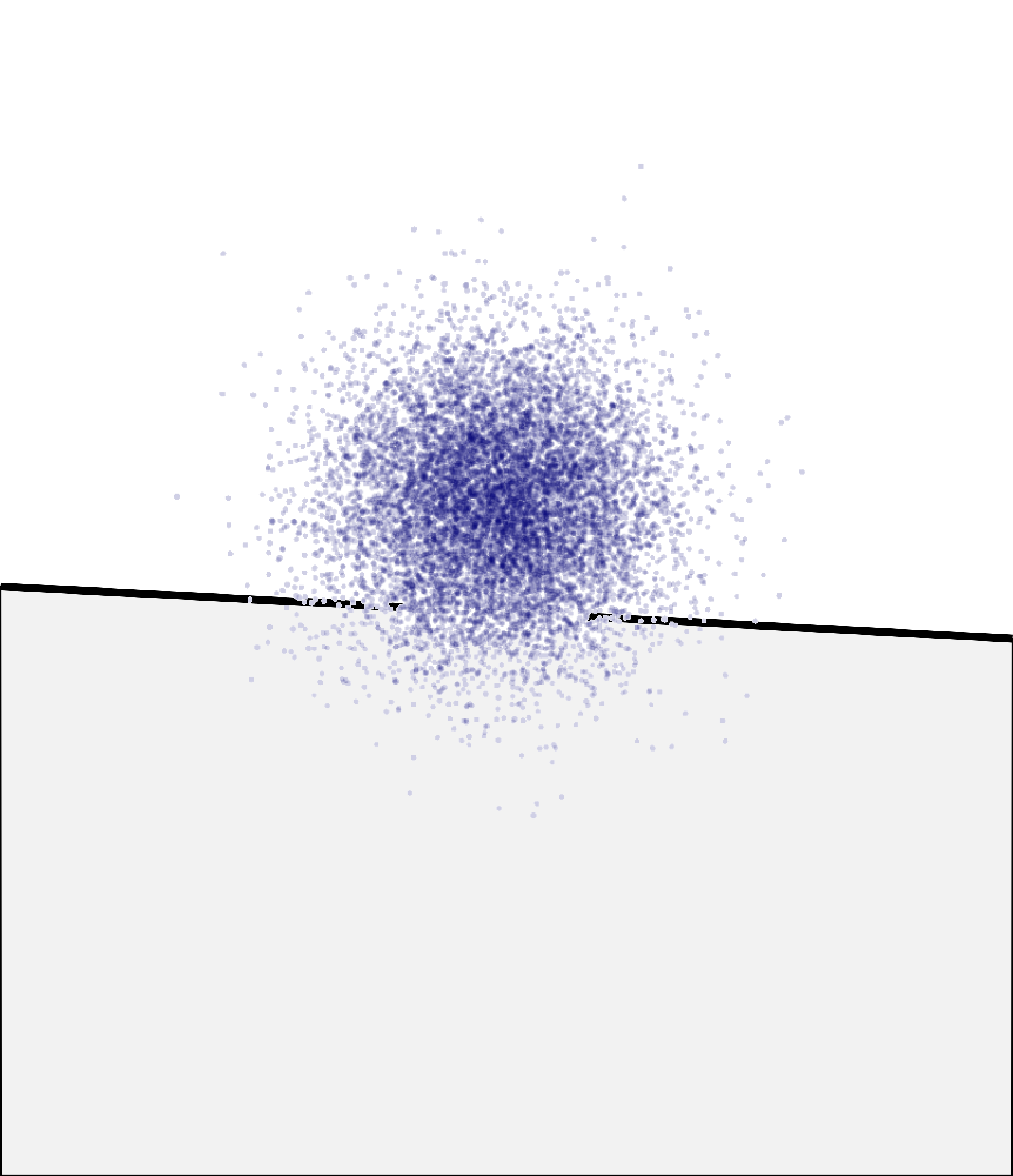}
	\includegraphics[width=0.240\columnwidth]{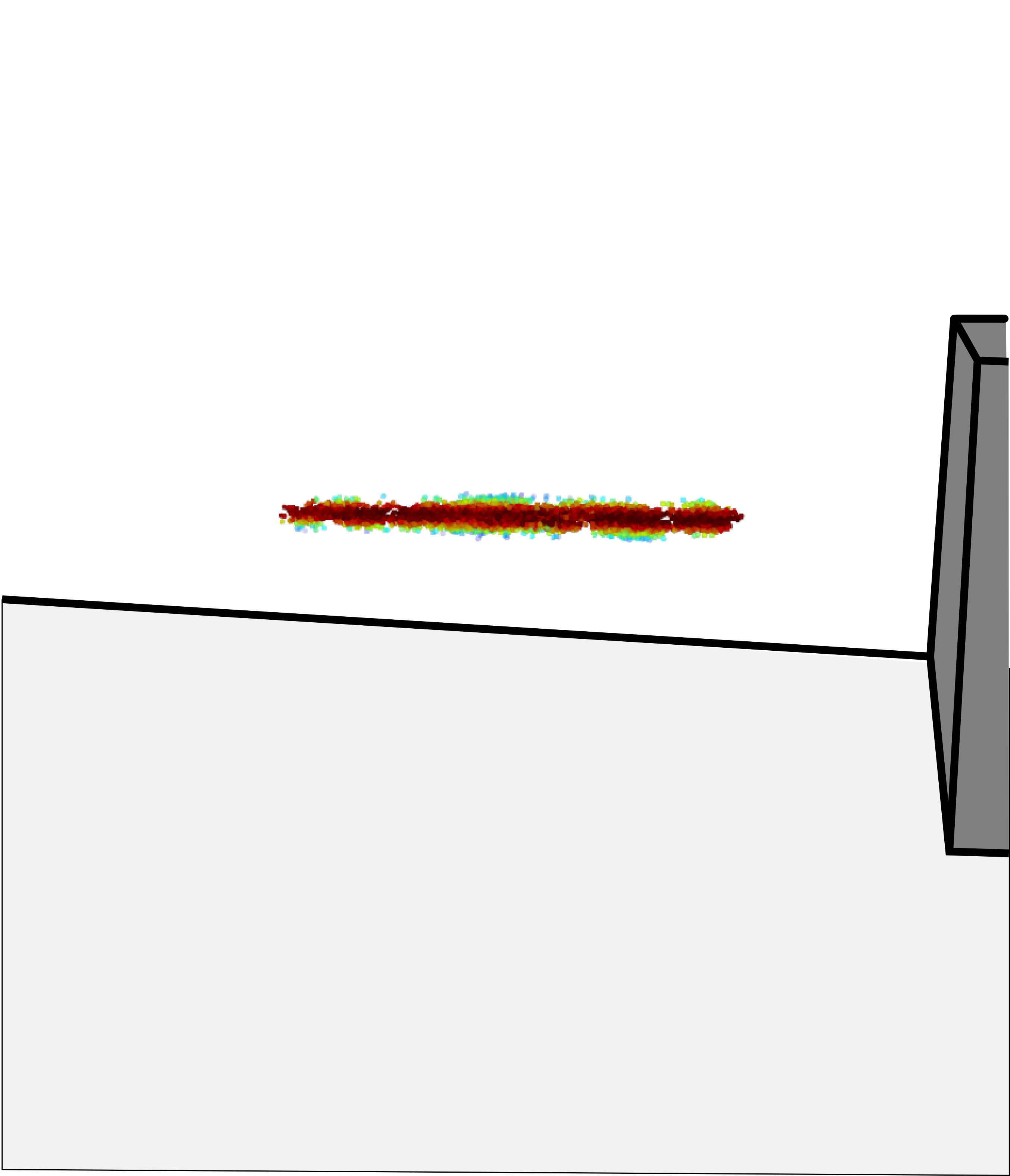}
	\includegraphics[width=0.240\columnwidth]{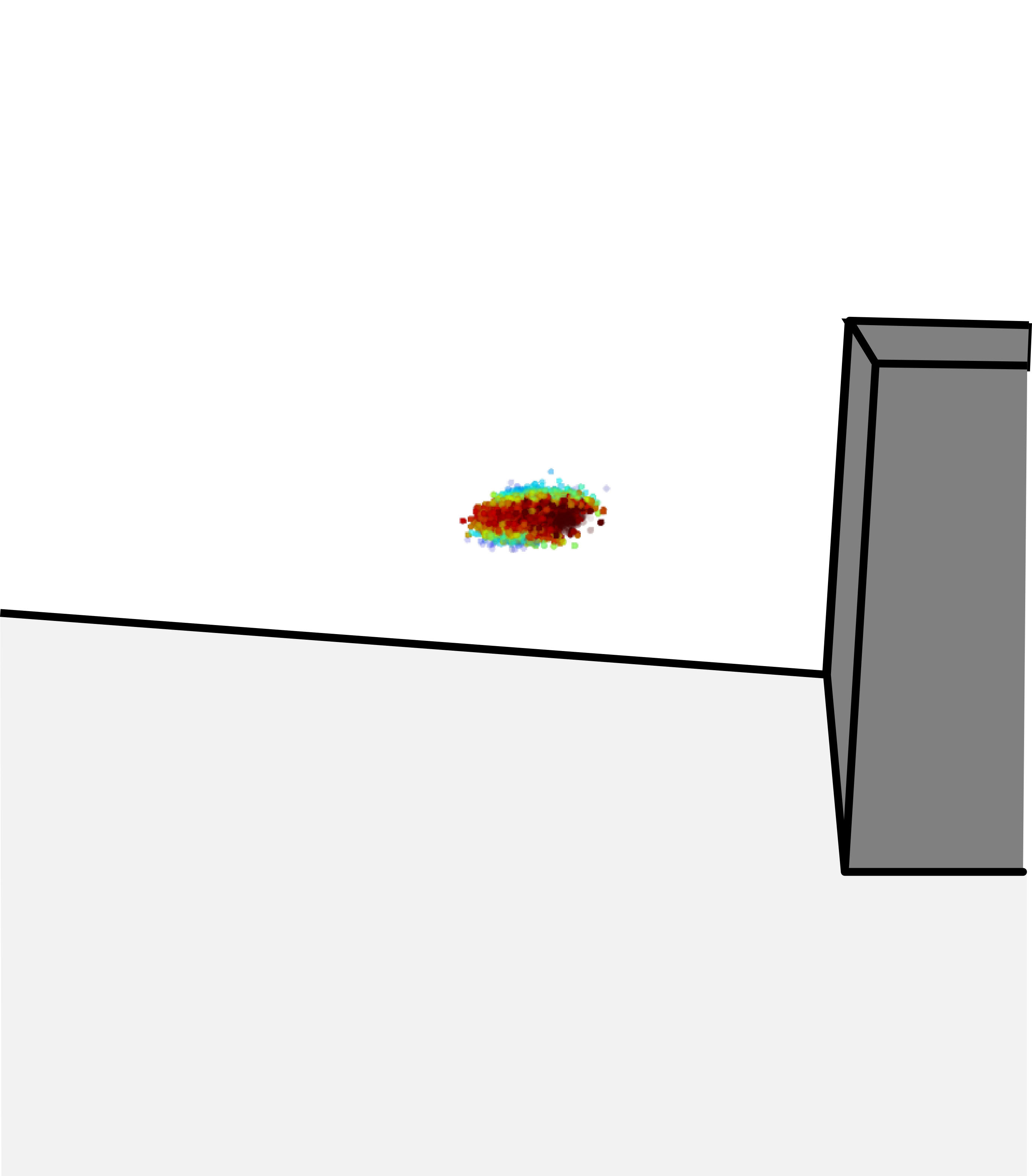}
	\includegraphics[width=0.240\columnwidth]{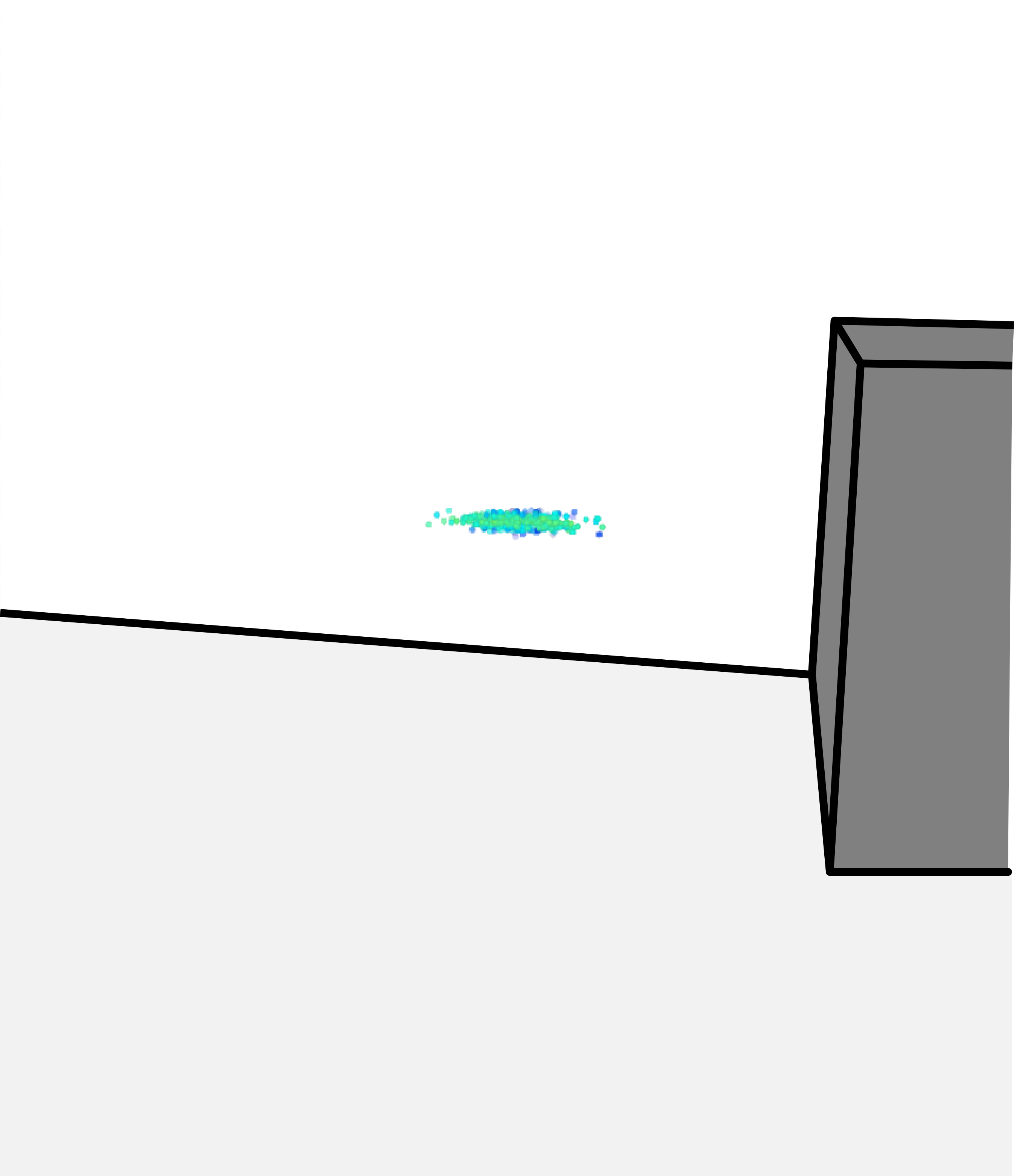}
	\caption{Experiment 2: Haptic probing experiment in 3-dimensions. The top row shows the
robot performing the experiment while the bottom row shows the particle
distribution and localization estimate. The particle set is colorized by normalized weight according to the jet colormap (i.e., dark blue = lowest weight, dark red = highest weight).  First, on the bottom left, the robot has an initial distribution with poses equally weighed. The robot makes a forward probe and then moves to the right. Now the particles are distributed as an ellipse with high uncertainty to the left and right of the robot. Then, the robot makes a probe to the right and touches an obstacle; the particle cloud collapses into a tight cluster. Since the robot is now localized, it is able to complete the task of pressing the button on the wall.}
	\label{fig:3d-probing}
\end{figure}

As shown in Figure~\ref{fig:3d-probing} the robot was able to correct its
localization and complete the task of touching the button. The initial probe to the front reduced uncertainty in the robot's $x$ and $z$ directions, which
reduces the particles to an ellipsoidal elongated along $y$. As the robot moves, uncertainty in the $x$ direction increases slightly. By touching the wall on the right, the robot re-localized in all three dimensions in much the same way as a human following a wall in the dark would. The re-localization allows the robot to press the button, demonstrating the generalization of our algorithm to 3D. This would enable a robot to localize by probing a known piece of mining machinery, allowing it to perform maintenance tasks. The final position error was: $[7.7, -3.7, -0.2]$ centimeters in the $x$, $y$ and $z$ directions.

\subsection{Experiment 3: Terrain Classification}
\label{sec:experiment-class}
\begin{figure}
  \centering
  \includegraphics[width=\columnwidth]{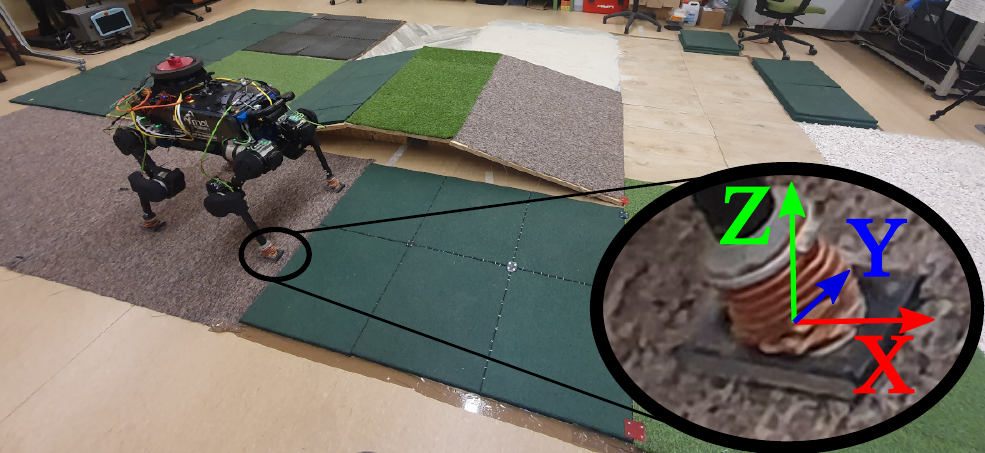}
  \caption[ANYmal haptic localization experiment]{Experiment 3: ANYmal with sensorized flat feet standing on the multiple terrain type course. Close up of the foot is provided. An IMU and fore/torque sensor is located in the sole with coordinate frame shown.}
\label{fig:anymal-terrain-class}
 \end{figure}

In the third experiment, we demonstrate the localization using terrain semantic information, which has been tested on a custom designed terrain course. Multiple \SI{1 x 1}{\meter} tiles of different terrain materials were placed on a \SI{3.5 x 7}{\meter} area. The course includes a \SI{20}{\centi\meter} high platform with two ramps with different terrain materials, as shown in Fig~\ref{fig:anymal-terrain-class}. The different terrain types used were gum, carpet, PVC, foam, sand, artificial
grass, ceramic and gravel. 

For the terrain classifier training, we gathered an additional dataset of the robot walking on the different patches that consists of 8773 steps. The dataset was split into 7018 training and 1775 testing samples.
The network was trained following the steps described in section~\ref{sec:training}.
We obtained a classification accuracy of \SI{94.12}{\percent}.
The mean and standard deviation of the accuracy was estimated from k-fold cross-validation to be \SI{94.03}{\percent} and 0.09, respectively.




\begin{table}
\centering
\begin{tabular}{ccccc}
\toprule
\multicolumn{5}{c}{\textbf{Mean Absolute Translation Error (ATE)
[\si{\meter}]}} \\
\toprule
\textbf{Trial} & \textbf{Dist. [\si{\meter}]} & \textbf{Time [\si{\second}]} &
 \textbf{HL-G} &
\textbf{HL-GC}\\
\toprule
1 & 191 & 1114 & 0.23  & \textbf{0.14} \\
\midrule
2 & 331 & 1850 & 0.25  & \textbf{0.11}  \\
\midrule
3 & 193 & 1090 & 0.21  & \textbf{0.18} \\
\bottomrule
\end{tabular}
 \caption{Experiment 3: HL-G = Haptic Localization with only geometry \cite{buchanan2020haptic}; HL-GC = Haptic Localization with both geometry and terrain class. }
\label{tab:rpe2}
\end{table}


The robot was equipped with sensorized feet from \cite{ethfeet} and autonomously walked between pre-programmed waypoints placed over the entire course, including several passes over the ramp. Large sections of the trajectory were only on the flat terrain tiles, forcing the algorithm to rely mostly on terrain classification for localization. Unlike Experiment 1, the robot was able to walk completely blind and no exteroception was used for footstep planning. A statically stable gait was used such that one foot was in the air at a time.

To demonstrate repeatability, we have performed three trials of this type, for a total distance traveled of more than \SI{0.7}{\kilo\meter} and \SI{1}{\hour}\SI{7}{\minute} duration. We compare results produced using HL-G (Geometry) and HL-GC (Geometry and Terrain Classification). As the majority of the terrain course is flat, there is not enough information for geometry only localization to be continuously accurate. Only when using terrain class information as well the robot can localize in all parts of the terrain course.

\begin{figure}
 \centering
 \includegraphics[width=\columnwidth]{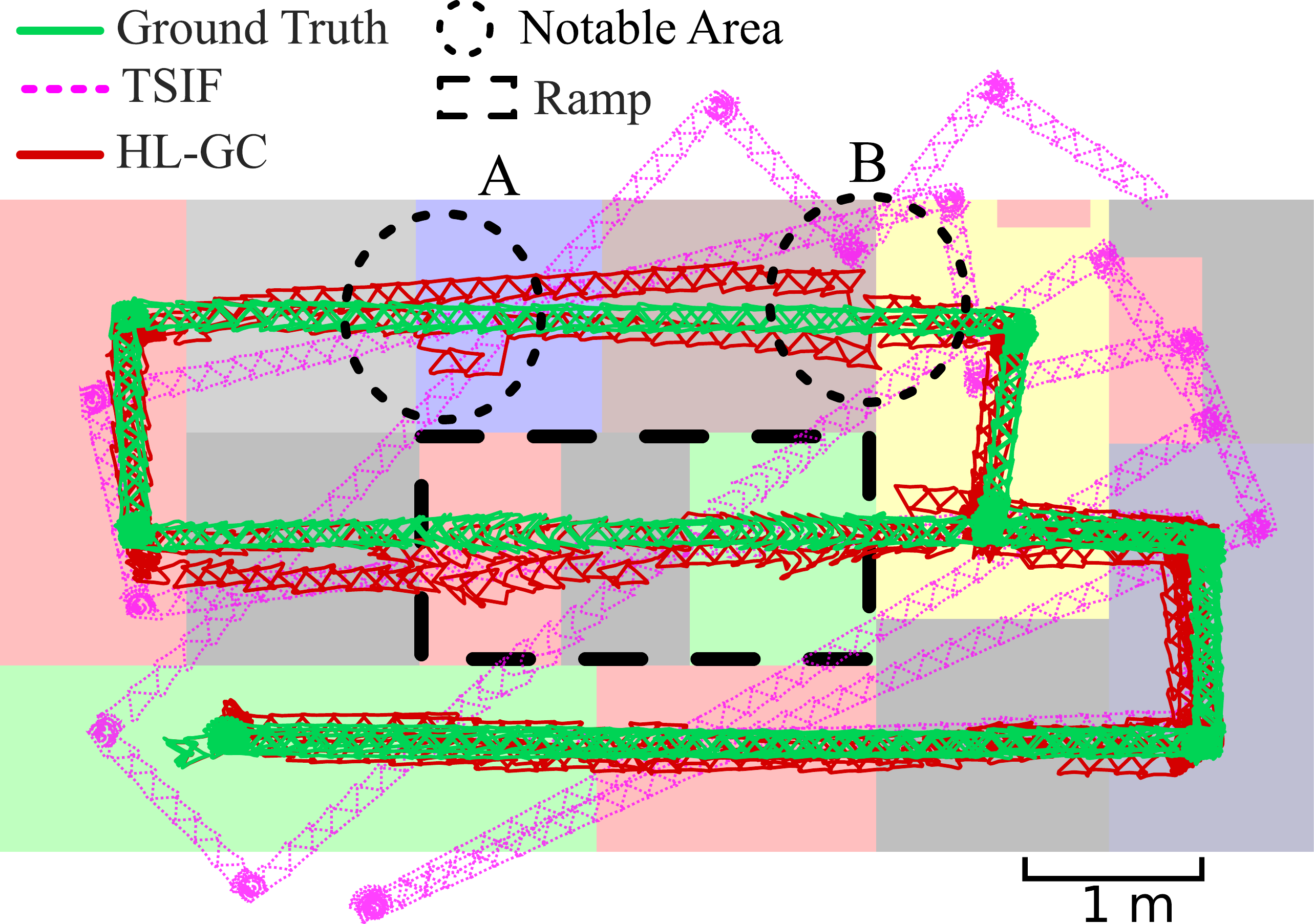}
 \caption{Experiment 3: Top down view of the state estimator (magenta), HL-GC estimated trajectory (red) and ground truth (dashed green) for trial 2. Trajectories are overlaid on terrain map with ramp shown in a dashed box. Two dotted circles show interesting areas in the trajectory. \textbf{A:} At a boundary crossing, the particle mean diverges from the ground truth. However, as the particle cloud nears the ramp, the geometric information gives higher likelihood to the particles in the center of the terrain. \textbf{B:} Another boundary crossing which in two separated crossings triggers good localization updates.}
 \label{fig:top-view-class}
\end{figure}

A summary of the experiments is presented in Table \ref{tab:rpe2}, where HL-GC shows an overall improvement between \SIrange{14}{56}{\percent} in the Absolute Translation Error (ATE) compared to HL-G. Using only the prior knowledge of the terrain geometry and class, the robot stayed localized in all the runs and bounded the linearly growing drift of the
state estimator. This can be seen in Figure~\ref{fig:top-view-class}, where the
estimated trajectory (in red) is able to stay near the ground truth trajectory (green).
In areas where there are large patches of the same material, such as the gravel
(dark blue) and ceramic (yellow), there is not enough information to localize in
the $xy$-plane and the localization drifts. When the robot crosses the
boundary into a new terrain type the localization is able to correct.

\begin{figure}
 \centering
 \includegraphics[width=\columnwidth]{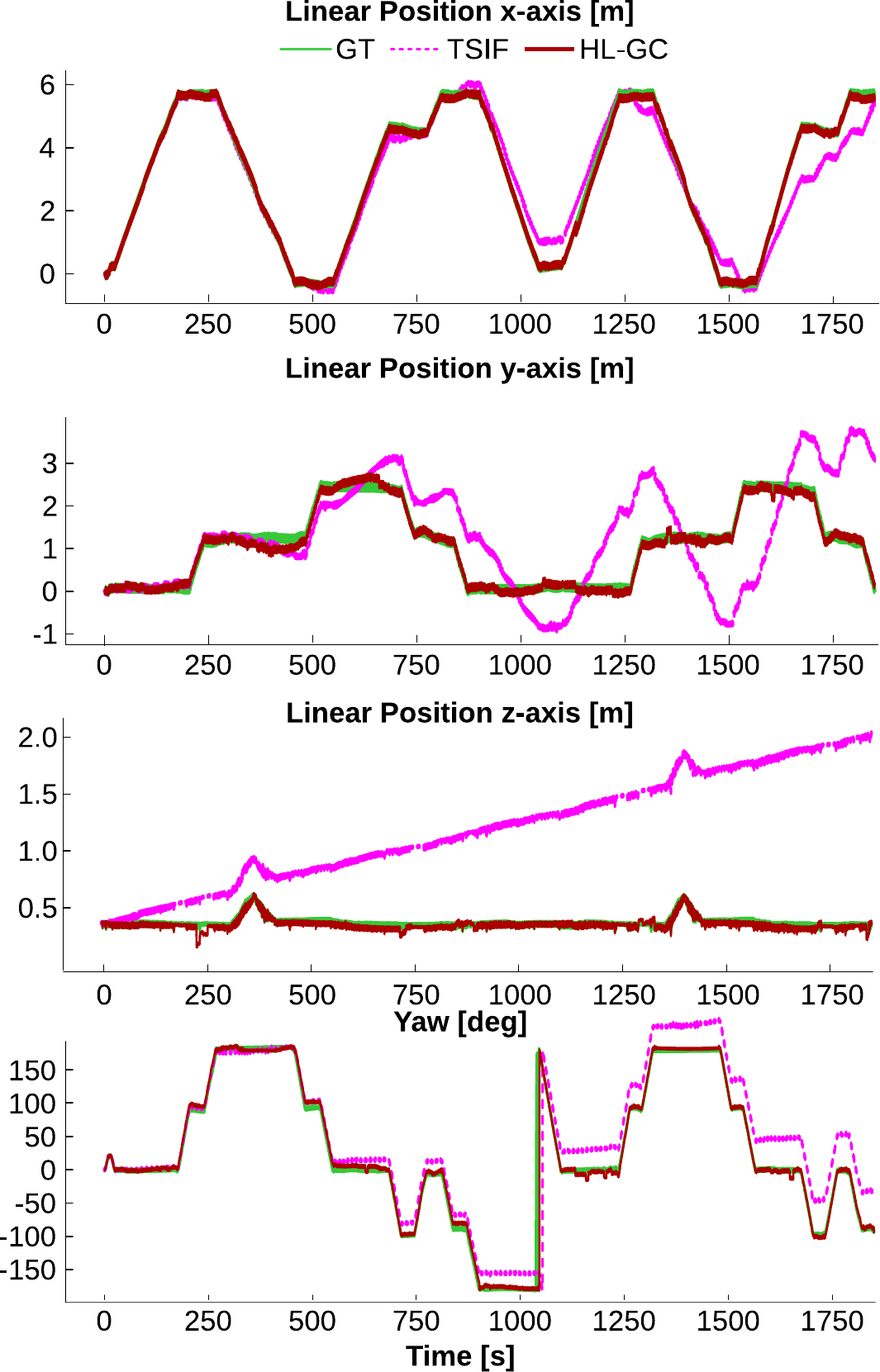}
 \caption{Experiment 3: Comparison between the estimated position from TSIF (dashed magenta)
and Haptic Localization (blue) against Ground Truth (green) for trial 2.}
 \label{fig:plots2}
\end{figure}

Figure~\ref{fig:plots2} shows in detail the estimator performance for each of
the three linear dimensions and yaw. As in Experiment 1, the error on the odometry filter (TSIF, magenta dashed line) of the robot is dominated by upward and yaw drift. This drift  is estimated and compensated for by the haptic localization (red solid line), allowing accurate tracking of the ground truth (green solid line) in all dimensions\footnote{A video showing all of these experiments is attached as supplementary material}. 

\section{Discussion}
\label{sec:discussion}

The results presented in Sections \ref{sec:experimental-2.5d} and \ref{sec:experiment-3d} demonstrate that terrain with a moderate degree of geometric complexity (such as Figure~\ref{fig:anymal-chevron}) already provides enough information to bound the uncertainty of the robot's location. The effectiveness of a purely geometric approach is obviously limited by the actual terrain morphology in a real world  situation, which would need to contain enough features such that all the DoF of the robot are constrained once the robot has touched them.

In the case where there is not enough geometric information, we have shown in Section \ref{sec:experiment-class} that terrain semantics can be used to localize. With sufficiently diverse terrain types (as shown in Figure \ref{fig:anymal-terrain-class}), boundary crossings from one terrain to another provide enough information to correct for drift in the $xy$-plane.

\subsection{Analysis of Particle Distribution on Geometric Terrain}
\begin{figure}
 \centering
 \includegraphics[width=0.240\columnwidth]{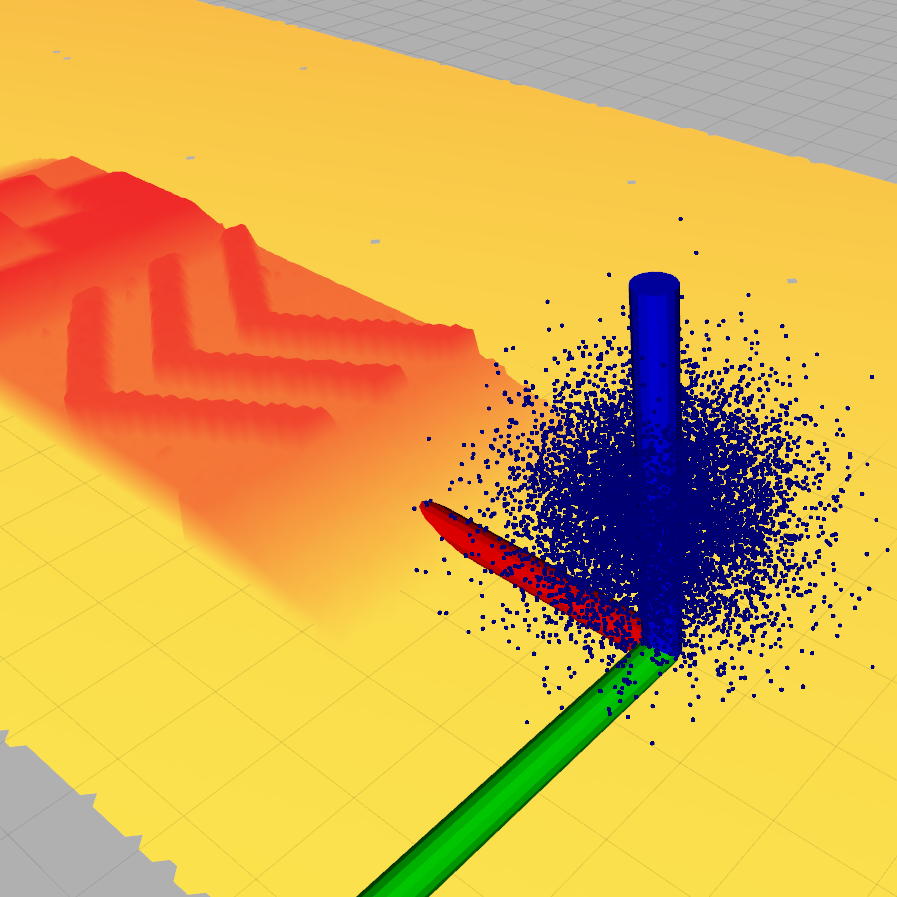}
 \includegraphics[width=0.240\columnwidth]{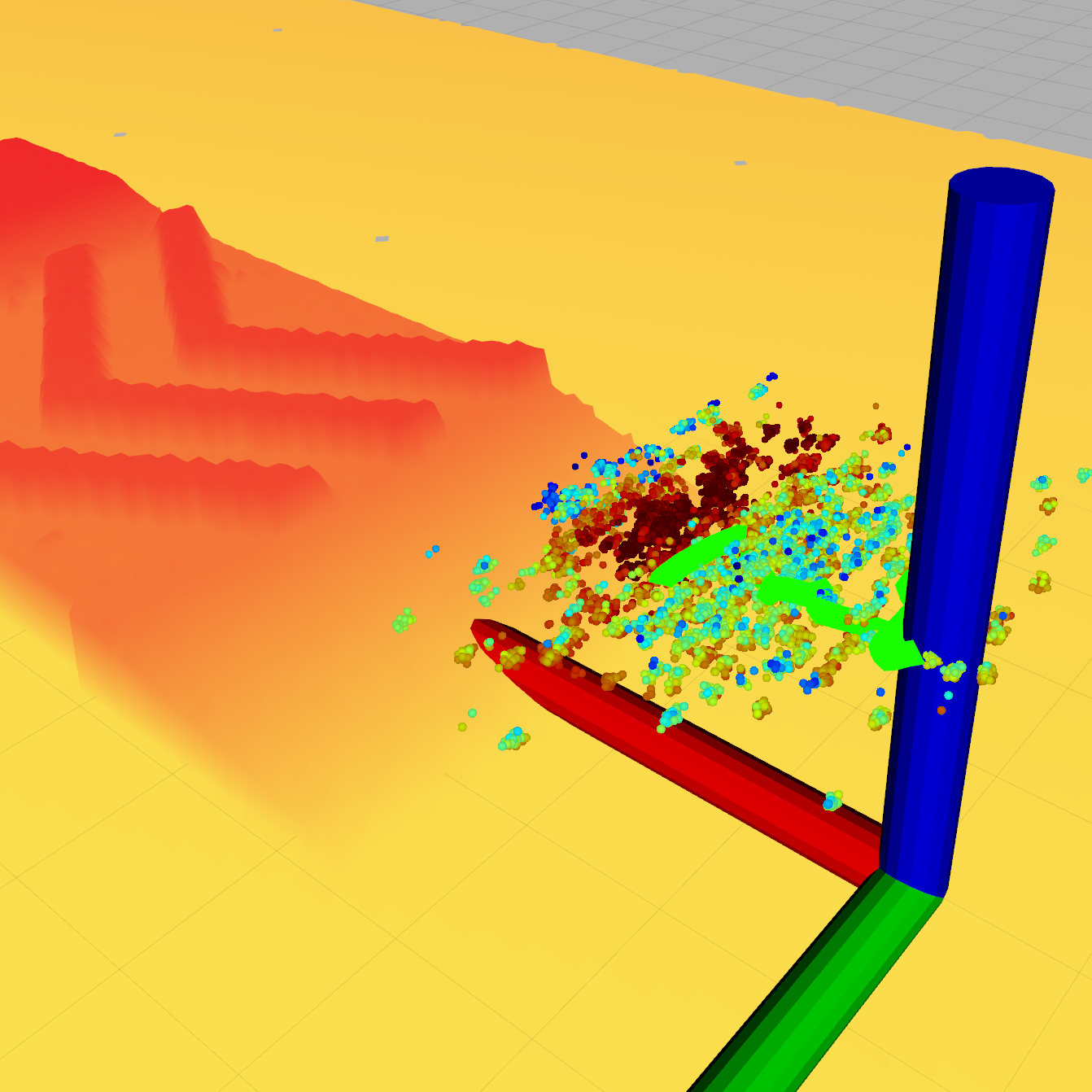}
 \includegraphics[width=0.240\columnwidth]{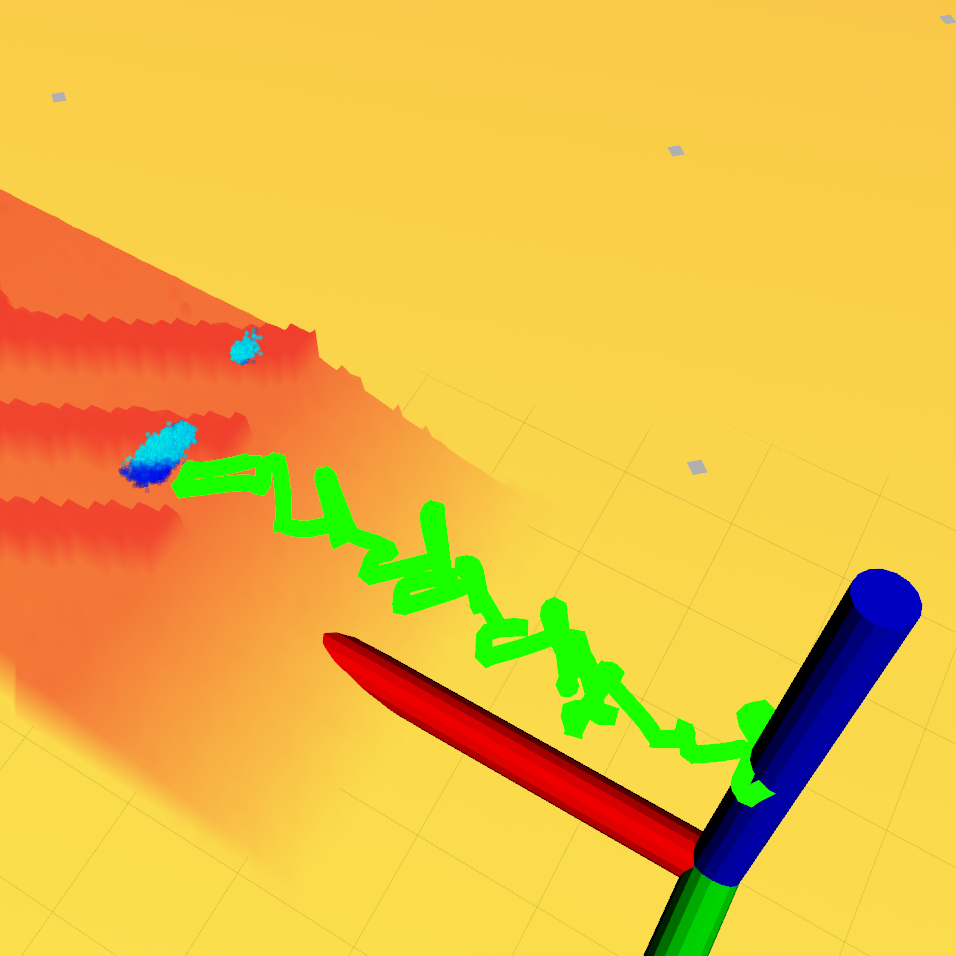}
 \includegraphics[width=0.240\columnwidth]{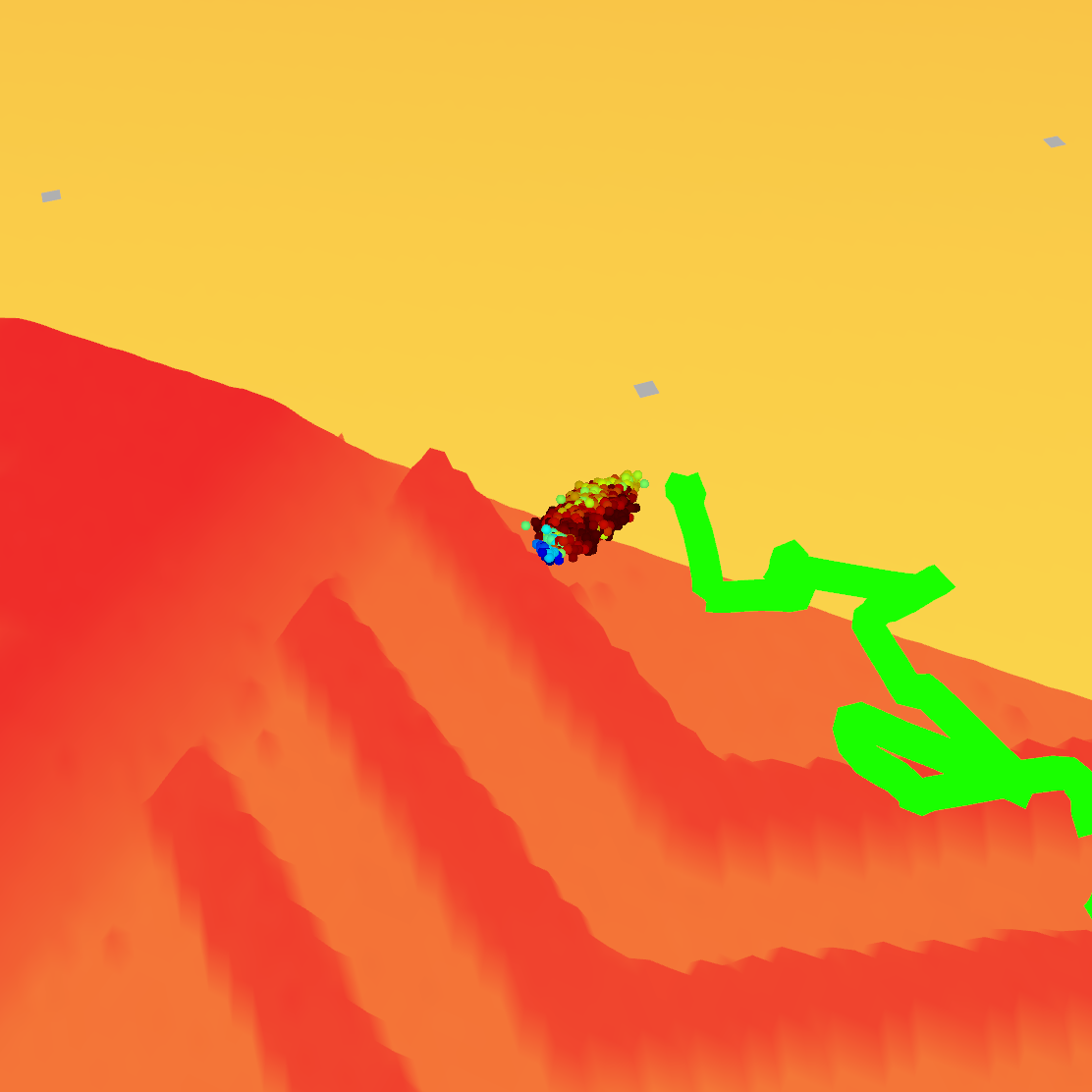}
 \newline
  \includegraphics[width=\columnwidth]{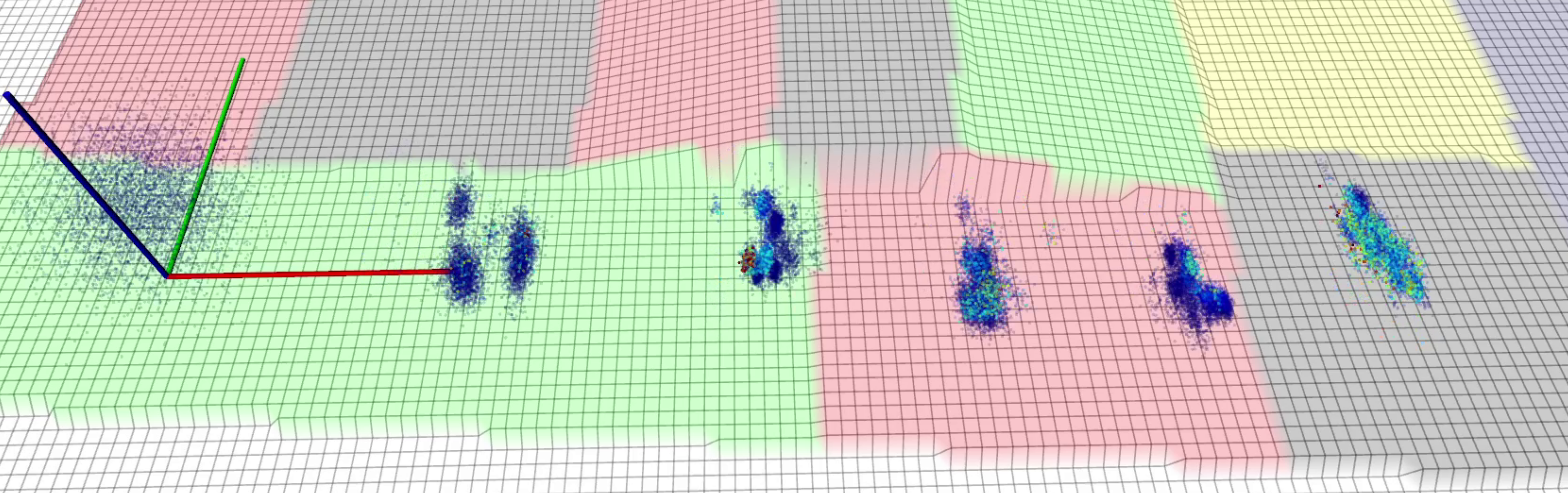}
 \caption{Evolution of two particle sets during experiments. The particle set is colorized by normalized weight according to the jet colormap (i.e., dark blue = lowest weight, dark red = highest weight). The green line indicates the ground truth trajectory. \textbf{Top (Experiment 1):} \emph{A)} At start, all the particles have the same weight and are normally distributed at the starting position. \emph{B)} After a few steps on the ramp, the robot pose is well estimated on $x$ and $z$ directions, but there is uncertainty on $y$. \emph{C)} When the robot approaches the chevron the particle set divides into two clusters, indicating two strong hypotheses as to the robot pose. \emph{D)} After a few more steps on the chevron, the robot is fully
localized and the particles are tightly clustered. \textbf{Bottom (Experiment 3):} As the robot walks from left to right, the particle cloud makes two terrain class transitions. As the robot crosses the first transition, the cloud becomes more narrow in the $x$ direction as error along this axis is corrected.}
 \label{fig:particles}
 \vspace{-10pt}
\end{figure}

Figure \ref{fig:particles} shows the evolution of the particles up to the first half of the terrain course for Experiment 1, Trial 2. As the robot walks through, the particle cluster becomes concentrated, indicating good convergence to the most likely robot pose. 


In the third subfigure, it can be noted how the probability distribution over the robot's pose follows a bimodal distribution, which is visible as two distinct clusters of particles. This situation justifies the use of particle filters, as they are able to model non-Gaussian distributions which can arise from a particular terrain morphology. In this
case, the bimodal distribution is caused by the two identical gaps in between the chevrons. In such situations, a weighted average of the particle cluster would lead to a poor approximation of the true pose distribution. Therefore, the particle evolution illustrated in Section \ref{sec:statistics} is crucial to reject such an update.

\subsection{Analysis of Particle Distribution on Terrain Class}
\begin{figure}
 \centering
 \includegraphics[width=\columnwidth]{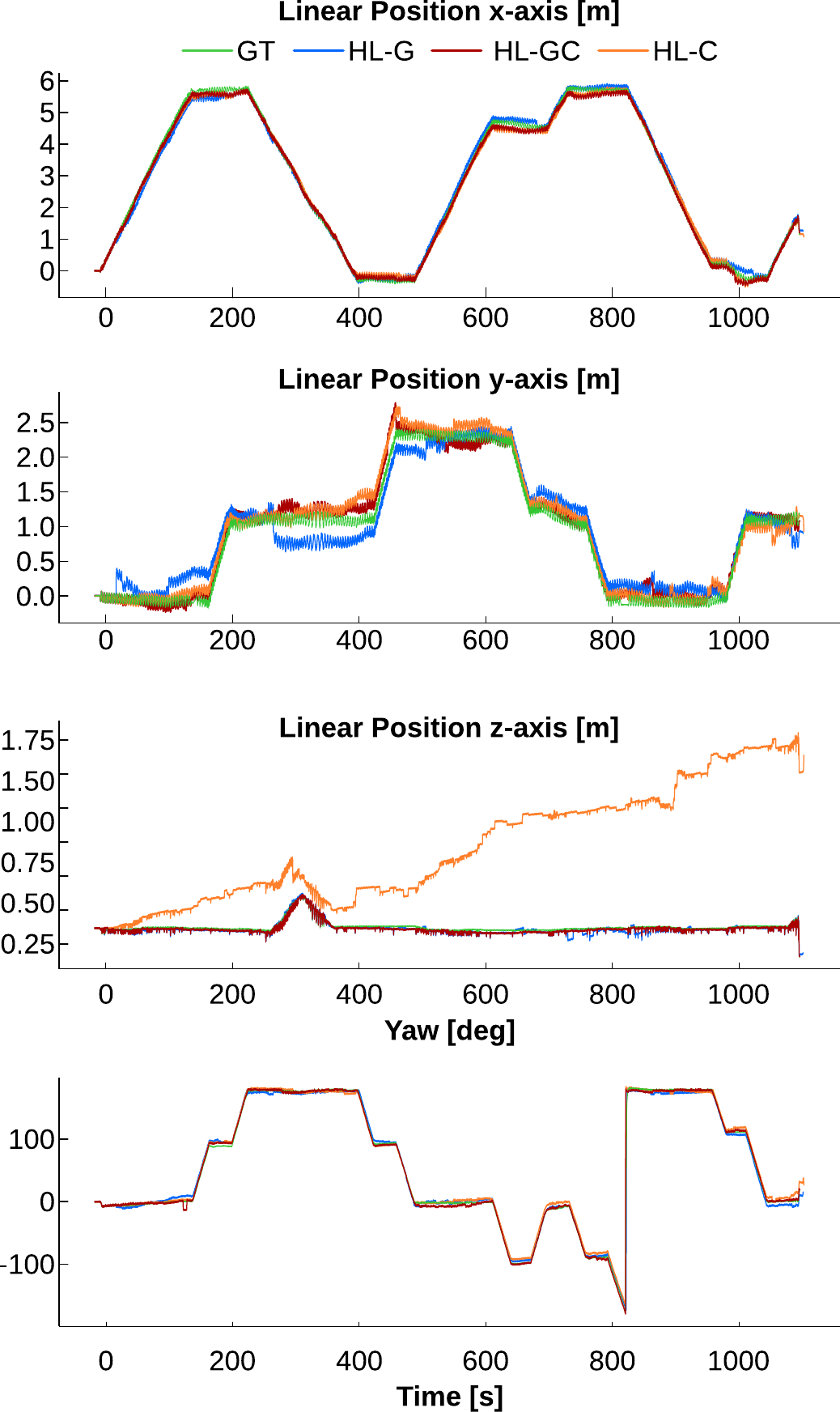}
 \caption{Experiment 3: Results from Trial 1. Here we show the difference in results when terrain class information and elevation are selectively removed. The median ATE for using only class data was 0.63\,m and for using only elevation data was 0.23\,m and for using both was 0.14\,m.}
 \label{fig:plots3}
\end{figure}

Figure \ref{fig:plots3} shows data from Experiment 3, Trial 1. We compare results from HL-G, HL-GC and HL-C (Haptic Localization using class data only). We can see that even with only class information, this method is able to keep localization error bounded in the $x$-$y$ plane (mean ATE for the class only trajectory was \SI{0.63}{\meter}). In the third subplot from the top, the class only trajectory drifts upward in a similar way to TSIF in Figure \ref{fig:plots2}. This is because of the absence of any measurement in the $z$, hence our method relies on the proprioceptive state estimator.

Further analysis of the effect of terrain class on localization is shown in Figure \ref{fig:particles}. Here, we show the evolution of particles from HL-GC in Experiment 3, Trial 3. The particles, which initially are normally distributed around the starting position, quickly converge along the $z$ axis as the floor elevation information is used. Once the boundary transition from green to red occurs, the particles correct for drift in the $x$ direction.

\section{Conclusion}
\label{sec:conclusion}
We have presented a haptic localization algorithm for quadrupedal robots based on Sequential Monte Carlo methods. The algorithm can fuse geometric information (in 2.5D or 3D) as well as terrain semantics to localize against a prior map. We have demonstrated that even using only geometric information, walking over a non-degenerate terrain course containing slopes and interested geometry can reduce localization error to \SI{10}{\centi\meter}. Our method also works if the robot probes vertical surfaces, measuring its environment in full 3D. Finally, we have shown how in areas of even sparser geometric information, terrain semantics can be used to augment this geometry.

The proposed approach demonstrated an average of \SI{20}{\centi\meter} position error over all areas of a terrain course with different terrain classes and geometries. The ability to localize purely proprioceptively is valuable for repetitive autonomous tasks in vision-denied conditions, such as inspections of sewage systems. This method could also serve as a backup localization system in case of sensor failure --- enabling a robot to complete its task and return to base.

\section{Acknowledgments}
This research has been conducted as part of the ANYbotics research community. It
was part funded by the EU H2020 Project THING (Grant ID 780883) and a Royal Society University
Research Fellowship (Fallon).

\bibliographystyle{spbasic}      
\bibliography{./library}
\end{document}